\documentclass[journal]{IEEEtran}
\IEEEoverridecommandlockouts
\usepackage{amsmath,amssymb,amsfonts}
\usepackage{amsthm} 
\usepackage{algorithmic}
\usepackage{graphicx}
\usepackage{textcomp}
\usepackage{xcolor}
\usepackage{ulem}
\usepackage[numbers,sort&compress]{natbib} %IEEE仕様でnatbibを召喚
\makeatletter
\renewcommand{\@biblabel}[1]{[#1]\hfill} %文献リストの表示を定義
\makeatother

\newcommand{\pdf}{p}
\newcommand{\partition}{\rho}
\newcommand{\vol}{\operatorname{vol}}
\newcommand{\E}{\mathbb{E}}
\newcommand{\ceil}[1]{\lceil #1 \rceil}
\newcommand{\supS}[1]{\sup_{x \in S} (#1)}
\newcommand{\infS}[1]{\inf_{x \in S} (#1)}
\newcommand{\bluetext}[1]{\textcolor{black}{#1}}
\usepackage{comment}

\theoremstyle{plain}
\newtheorem{thm}{Theorem}
\newtheorem*{thm*}{Theorem}

\theoremstyle{definition}
\newtheorem{dfn}{Definition}

\theoremstyle{plain}
\newtheorem{cor}{Corollary}
\newtheorem*{cor*}{Corollary}

\theoremstyle{remark}
\newtheorem{remark}{Remark}

\newtheorem{lem}{Lemma}
\newtheorem*{lem*}{Lemma}

\newtheorem{exa}{Example}
\newtheorem*{exa*}{Example}

\newtheorem{proposition}{Proposition}

\def\BibTeX{{\rm B\kern-.05em{\sc i\kern-.025em b}\kern-.08em
    T\kern-.1667em\lower.7ex\hbox{E}\kern-.125emX}}

\begin{document}

\title{Normalized Maximum Likelihood Code-Length on Riemannian Data Spaces\\}

\author{Kota~Fukuzawa, Atsushi~Suzuki, and~Kenji~Yamanishi,~\IEEEmembership{Senior Member,~IEEE}% 
\thanks{Manuscript submitted February 9, 2026. This work was partially supported by JST KAKENHI Grant Number 24H00703.}% 
\thanks{K. Fukuzawa was with The University of Tokyo, Tokyo, Japan, where most of the work presented in this paper was carried out. He is now with NTT, Inc., Tokyo, Japan (e-mail: k.fukuzawa@ntt.com).}% 
\thanks{A. Suzuki is with The University of Hong Kong, Pokfulam, Hong Kong (e-mail: atsushi.suzuki.rd@outlook.com). Corresponding author.}% 
\thanks{K. Yamanishi is with The University of Tokyo, Tokyo, Japan (e-mail: yamanishi@g.ecc.u-tokyo.ac.jp).}% 
}

\maketitle

\begin{abstract}
In recent years, with the large-scale expansion of graph data, there has been an increased focus on Riemannian data spaces other than Euclidean space. In particular, the development of hyperbolic spaces has been remarkable, and they have high expressive power for graph data with hierarchical structures. Normalized Maximum Likelihood (NML) is employed in regret minimization and model selection. However, existing formulations of NML have been developed primarily in Euclidean spaces and are inherently dependent on the choice of coordinate systems, making it non-trivial to extend NML to Riemannian manifolds. In this study, we define a new NML that reflects the geometric structure of Riemannian manifolds, called the Riemannian manifold NML (Rm-NML). This Rm-NML is invariant under coordinate transformations and coincides with the conventional NML under the natural parameterization in Euclidean space. We extend existing computational techniques for NML to the setting of Riemannian manifolds. Furthermore, we derive a method to simplify the computation of Rm-NML on Riemannian symmetric spaces, which encompass data spaces of growing interest such as hyperbolic spaces. To illustrate the practical application of our proposed method, we explicitly computed the Rm-NML for normal distributions on hyperbolic spaces.

\end{abstract}


\section{Introduction}
\IEEEPARstart{W}{ith} the recent increase in the scale of graph data, Riemannian data spaces other than Euclidean spaces are attracting attention as latent spaces suitable for graph embedding \cite{zhu2022spherical, nickel2017poincare}. For example, hyperbolic spaces have been demonstrated to possess high expressive power for graph data with hierarchical structures \cite{krioukov2010hyperbolic}. Spherical spaces are particularly effective in representing graph data with cyclic structures \cite{gu2018learning}. Notably, research on hyperbolic spaces has been particularly remarkable \cite{krioukov2010hyperbolic}. Specifically, in the field of representation learning, methods that embed hierarchical structures into hyperbolic space have successfully represented such relationships using significantly lower-dimensional space compared to conventional methods based on Euclidean space, while preserving the essential relational information \cite{nickel2017poincare}. By assuming a probability distribution over the low-dimensional latent space, which provides a compact and essential representation of data sequences, one can obtain an even more concise and informative characterization of the underlying data structure. When estimating the probabilistic structure behind data, techniques such as model selection and regret analysis are typically employed. Normalized Maximum Likelihood (NML) is known to be a useful approach in these contexts, as it achieves the minimax regret and is widely applied in data analysis based on code-length \cite{shtar1987universal}. Once NML-based encoding is realized, the application of the Minimum Description Length (MDL) principle becomes feasible \cite{yamanishi2023learning}. For point sequences in Euclidean space, the NML code-length is known to be optimal in terms of minimizing code-length regret \cite{shtar1987universal}.

Furthermore, detailed methods for calculating the NML code-length have been extensively studied \cite{rissanen2002fisher,rissanen2002mdl, suzuki2018exact, suzuki2021fourier, yamanishi1996randomized, roos2008monte, rissanen2012optimal}. These computational methods are theoretically significant as they provide the regret-minimizing code-length. However, while cases where the parameter space forms a Riemannian manifold have been extensively studied in the field of information geometry \cite{amari2000methods}, situations in which the data space itself is a non-Euclidean Riemannian manifold, such as hyperbolic space, pose fundamental challenges to the above framework. There are two non-trivial challenges in extending NML to Riemannian manifolds: how to define it and how to compute it. In what follows, we address both of these issues in detail. We first outline the challenges associated with defining NML on Riemannian manifolds.
\begin{enumerate}
\item For a continuous probability density function $\pdf(x)$, the quantity $-\log \pdf(x)$ is often interpreted as a code-length via a discretization of the space in which $x$ is embedded. However, in the case of data spaces endowed with a Riemannian manifold structure, the discretization process associated with coding is not straightforward. In Euclidean space, for instance, discretization using lattices is possible. On such a discretization, the negative log-likelihood can be interpreted—up to an additive constant determined by the discretization width—as a code-length, particularly when discretizing per unit volume. However, in the case of a Riemannian manifold, no lattice exists that uniformly reflects its intrinsic distance structure.
\item In Euclidean space, the canonical Cartesian coordinate system exists, and the negative logarithm of a probability density function defined on this system can be interpreted as a code-length. However, on a general Riemannian manifold, no such canonical coordinate system exists. Since probability density functions are coordinate-dependent, interpreting the negative log-density as a code-length introduces ambiguity: the resulting code-length depends on the choice of coordinates and is therefore not uniquely defined.
\item Without a well-defined notion of code-length, the NML code-length cannot be meaningfully formulated on a general Riemannian manifold.
\end{enumerate}

Even if the above issues were resolved, making such a formulation computationally feasible would remain a significant challenge. In practice, computing the NML code-length is challenging even in Euclidean spaces. The NML code-length is defined as the code-length under the NML distribution, which consists of the maximum likelihood obtained using the maximum likelihood estimator (MLE) of the parameters from the data sequence, and a normalization term over the space of all possible data sequences \cite{shtar1987universal}. This normalization term, known as the parametric complexity (PC), is notoriously difficult to compute. While several methods have been developed to approximate the PC in Euclidean spaces using asymptotic analysis and related techniques \cite{rissanen2002fisher}, it remains non-trivial whether these methods can be extended to Riemannian manifolds, including non-Euclidean spaces. 

For Euclidean spaces, several methods for calculating the PC have been proposed in previous studies. Rissanen (1996) derived an asymptotic approximation formula for the PC based on the asymptotic normality of the MLE \cite{rissanen2002fisher}. Subsequently, Rissanen (2000, 2012) developed a method for calculating the value of PC by partitioning the data sequence space according to the values of the maximum likelihood estimates of the parameters \cite{rissanen2002mdl, rissanen2012optimal}. Mathematical justification of the method for continuous data space cases was recently given \cite{suzuki2024foundation}, based on geometric measure theory. Additionally, Suzuki \& Yamanishi (2018, 2021) proposed a Fourier method \cite{suzuki2018exact, suzuki2021fourier}. Furthermore, Yamanishi (1996) and Roos (2008) proposed a Monte Carlo method for estimating the PC value using sampling approximation \cite{yamanishi1996randomized, roos2008monte}. The aforementioned PC calculation methods have been summarized by Yamanishi (2023) \cite{yamanishi2023learning}. However, most of the existing computational techniques and methodologies have been developed exclusively for probabilistic models defined on Euclidean data spaces. To the best of our knowledge, there exists no prior work that addresses the computation of the NML code-length on Riemannian manifolds.

This study extends the notion of code-length to cases where the data space is a Riemannian manifold and defines the Riemannian manifold Normalized Maximum Likelihood (Rm-NML) code-length, which is optimal in terms of regret with respect to such a code-length. Furthermore, we provide a method for computing the Rm-NML code-length.

The code-length we consider on Riemannian manifolds is defined as the negative logarithm of a probability density function with respect to the base measure induced by the volume element. This corresponds to the code-length obtained by discretizing with unit volume, disregarding multiplicative factors due to discretization methods. To the best of our knowledge, this is the first formulation of its kind. The Rm-NML we define minimizes the regret with respect to this geometrically meaningful notion of code-length, making it the first code-length with such a property.

We also present an asymptotic approximation method for computing the Rm-NML code-length. Specifically, we generalize the method proposed by Rissanen (1996) for Euclidean data spaces to the setting of Riemannian manifolds \cite{rissanen2002fisher}. Our method avoids integration over the data space’s Cartesian product and provides uniform asymptotic accuracy (with differences of order o(1)) over all data sequences, making it practical for applications. Importantly, the sufficient conditions for applying this method can be verified in any coordinate system, which is crucial because standard coordinate systems generally do not exist on Riemannian manifolds. In fact, we prove that the sufficient conditions are equivalent regardless of the choice of coordinates.

Moreover, we present a simplified computation method for the Rm-NML code-length in cases where the probability density function depends on the Riemannian data point only through the distance function to a parameter. The motivation for considering non-Euclidean spaces, such as hyperbolic space, as data domains lies in the fact that their intrinsic distance structures are well-suited for representing certain types of data. When constructing probabilistic models on such spaces, it is therefore natural to incorporate this underlying geometric structure into the formulation of the probabilistic models. This is a special case of the general method and inherits all its desirable properties. Since most practical applications of Riemannian manifolds focus on their coordinate-invariant distance structure, this simplified computation is widely applicable. As a concrete example, we demonstrate how the Rm-NML code-length can be computed for the hyperbolic Gaussian distribution, a canonical probabilistic model on hyperbolic space.

\bluetext{One of the main contributions of this study is to enable the construction of a code that functions as stochastic complexity (SC) on Riemannian data spaces. SC is understood as a generalization of Shannon information to settings in which the underlying probabilistic model is unknown. When the standard NML is applied to Riemannian data spaces, the Kraft inequality may fail to hold depending on the choice of coordinate system. This issue arises because, on Riemannian manifolds, the standard NML code-length is not correctly defined as a code-length assigned to a partition into unit-volume elements. As a result, this leads to difficulties in applications to information theory. The Rm-NML code-length proposed in this work is coordinate-invariant and is defined with respect to the volume element. Consequently, Rm-NML enables the construction of Huffman codes in a manner that is independent of the choice of coordinate system.}
\subsection{Related research}
The NML addressed in this study plays a central role in the MDL principle. The MDL principle is a universal learning strategy grounded in the concept of data compression \cite{rissanen1978modeling, rissanen2012optimal}. Its theoretical validity has been established from various perspectives, consistency as in Rissanen (1998) \cite{rissanen1998stochastic}, convergence rate as in Barron and Cover (1991) \cite{barron2002minimum}, and sample complexity in stochastic probably approximately correct learning as in Yamanishi (1992) \cite{yamanishi1992learning}. More detailed theoretical foundations of the MDL principle are discussed in Gr\"{u}nwald (2007) \cite{grunwald2007minimum}. According to the MDL principle, the best model is the one that minimizes the total description length of the data and the model, as justified from an information-theoretic standpoint. \bluetext{This total description length is characterized by the NML code-length and is established as the SC} \cite{rissanen2012optimal}. The NML code-length is employed because it provides the unique solution to the minimax regret problem, as shown by Shtar'kov \cite{shtar1987universal}. Furthermore, Rissanen (1996) derived an asymptotic approximation for the computation of the NML code-length \cite{rissanen2002fisher}. \bluetext{One of the representative applications of NML is model selection based on the MDL principle. In this context, the NML code-length serves as a model selection criterion, and the probabilistic model with the shortest NML code-length is selected for a given observed data sequence. Model selection based on the NML code-length has been widely applied, and numerous applications have been reported to date \cite{tabus2003classification,myung2006model, roos2008bayesian, schmidt2010estimating, zhang2012model, hirai2013efficient, staniczenko2014selecting, suzuki2016structure, yamanishi2019decomposed, kellen2020selecting, yamanishi2023detecting, kontkanen2005mdl}}. The MDL principle and NML are comprehensively summarized in Yamanishi (2023) \cite{yamanishi2023learning}.

In recent years, data with non-Euclidean structures has become increasingly common \cite{fletcher2004principal, dryden2016statistical}, leading to growing interest in Riemannian data spaces \cite{nickel2017poincare, falorsi2019reparameterizing}. These spaces have garnered particular attention in the context of graph representation learning. For example, hyperbolic spaces have demonstrated remarkable success as embedding spaces for hierarchical graph data \cite{nickel2017poincare}, while spherical manifolds are known to be well-suited for representing cyclic structures in graphs \cite{zhu2022spherical}. Beyond these, other Riemannian manifolds such as tori and Möbius strips have also been explored as data domains \cite{ebisu2018toruse, chen2021mobiuse}. More recently, deep learning has been extended to Riemannian manifolds, enabling geometric learning directly on curved spaces \cite{ganea2018hyperbolic}. The theoretical foundation for optimization on Riemannian manifolds has been systematically presented in Absil et al. (2008) \cite{absil2008optimization}. Furthermore, various algorithms have been generalized to Riemannian settings: for instance, Bonnabel (2013) proposed a stochastic gradient descent method on Riemannian manifolds \cite{bonnabel2013stochastic}, and also extended the expectation-maximization (EM) algorithm to such spaces \cite{said2017riemannian}. Despite this growing body of research, there have been no studies that extend NML—which plays a central role in optimal model selection and regret analysis—to Riemannian data spaces.

As prior research on hyperbolic spaces and the MDL principle, there is a study by Yuki, Ike, and Yamanishi (2021) on dimension selection in hyperbolic spaces \cite{yuki2023dimensionality}. This study employs the Decomposed Normalized Maximum Likelihood Code-length (DNML), a model selection criterion based on the MDL principle for latent variable models \cite{yamanishi2019decomposed}. In DNML, the computation of the NML for the observed variables and that for the latent variables are performed separately. This study is the first to adopt non-Euclidean spaces as data spaces and address the MDL principle. Its performance is validated through experiments. However, a key difference from the present study is that Yuki, Ike, and Yamanishi (2021) adopt wrapped normal distributions (WNDs) as the probability model on hyperbolic spaces \cite{nagano2019wrapped}. WNDs are normal distributions defined on the tangent space of a manifold, analogous to those on Euclidean spaces, and are constructed by mapping the normal distribution onto the manifold via an exponential map.
Therefore, the NML is computed according to its standard definition \cite{shtar1987universal}, and there are no computational issues. In this study, we consider probability density functions that are defined directly on Riemannian manifolds, without relying on an embedding into or transformation from Euclidean space. As prior work dealing with manifold data spaces and the MDL principle, Yuki, Suzuki \& Yamanishi (2023) conducted research on dimension and curvature selection using DNML \cite{yamanishi2019decomposed}. This study also adopts WNDs as the probabilistic model. 

\bluetext{This study extends the normalized maximum likelihood (NML) code-length so that it can be applied to data spaces that are Riemannian manifolds. In the literature on the MDL principle and NML, there exist studies that interpret the parameter space as a statistical manifold grounded in information geometry. In particular, Balasubramanian (2005) showed that by adopting the Jeffreys prior, which corresponds to a uniform distribution on a statistical manifold from the viewpoint of information geometry, the NML code-length can be derived from a Bayesian perspective and from the geometric structure of parametric model families \cite{balasubramanian2005mdl}. This work provides a unified explanation of Occam’s razor, MDL, and Bayesian inference, and demonstrates that the quality of a model is determined not merely by the number of parameters but by its geometric robustness and expressive power.
A major difference between that study and the present work lies in the treatment of the data space. While Balasubramanian (2005) views the parameter space as a statistical manifold in the derivation of the general NML, the resulting NML is defined for data in Euclidean spaces. In contrast, the present study proposes Rm-NML, an extension of NML that operates on data spaces endowed with Riemannian manifold structures.
As another line of related work that treats the parameter space as a statistical manifold based on information geometry, Sun and Nielsen (2025) evaluated the complexity of deep neural networks under the MDL principle \cite{sun2019geometric}. In their analysis, the parameter space is modeled as a manifold using the framework of singular semi-Riemannian geometry. However, their study also considers data spaces that are Euclidean, which constitutes a key distinction from the present work. }\bluetext{As described above, there exist studies on NML that, based on information geometry, regard the parameter space as a Riemannian manifold. In contrast, in this study we treat both the data space and the parameter space as Riemannian manifolds. The proposed Rm-NML is a concept defined on Riemannian-manifold data spaces and yields a quantity that is invariant with respect to the choice of coordinate system on the data space. An important point here is that coordinate transformations in the parameter space and those in the data space are mathematically distinct, and independent coordinate transformations can be applied to each. All concepts introduced in this paper possess coordinate invariance even when such independent transformations are applied. In this way, this study resolves, in a manner consistent with Riemannian geometry, the issue of coordinate invariance in the data space, which has not been addressed in previous work.}

\subsection{Organization}
The paper is organized as follows. In Section \ref{Sec2}, we introduce Riemannian manifolds and the conventional NML. In Section \ref{Sec3}, we discuss the coordinate dependence of the conventional NML, which poses a major obstacle to its extension to Riemannian manifolds. Section \ref{Sec4} presents the definition of Rm-NML, an extension of NML to Riemannian manifolds. In Section \ref{Sec5}, we summarize the method for computing the PC on Riemannian manifolds, as derived in this study. Section \ref{Sec6} presents the computation results of Rm-NML for Gaussian distributions on hyperbolic space, based on the proposed definition and computational method. Finally, Section \ref{Sec7} concludes the paper. The appendix contains proofs of theorems and detailed derivations of various results.

\section{Preliminary}
\label{Sec2}
%\label{Preliminary}
\subsection{Riemannian manifold}
The definition of a Riemannian manifold is given as follows. 
\begin{dfn}[Riemannian manifold]
Let $\mathcal{M}$ be a differentiable manifold. Suppose that each tangent space $T_{p}\mathcal{M}$ at point $p\in\mathcal{M}$ is equipped with an inner product, given as a positive-definite symmetric bilinear form as follows.
\begin{equation}
    g_p : T_{p}\mathcal{M} \times T_{p}\mathcal{M} \to \mathbb{R}.
\end{equation}
Let $(U, (x_1, \cdots, x_D))$ be a local coordinate chart around point $p$, and define the following at point $p$.
\begin{equation}
\label{R-metric}
    g_{ij}(p)=g_p\left(\left(\frac{\partial}{\partial x_i}\right)_{p}, \left(\frac{\partial}{\partial x_j}\right)_{p}\right).
\end{equation}
If the components $g_{ij}$ are of class $C^{\infty}$ in a neighborhood of each point in $\mathcal{M}$, then the assignment of an inner product $g_p$ to each point $p\in\mathcal{M}$ defines a Riemannian metric $g$ on $\mathcal{M}$. In this case, a differentiable manifold equipped with a Riemannian metric is called a Riemannian manifold. As an equivalent definition to Equation (\ref{R-metric}), the metric tensor can be expressed using the dual basis $\{dx^1, \cdots, dx^D\}$ of the cotangent bundle as follows.
\begin{equation}
    g=\sum_{i, j} g_{i j} \mathrm{~d} x_i \otimes \mathrm{~d} x_j.
\end{equation}
\end{dfn}
As a concrete example of a Riemannian manifold, we present the definition of hyperbolic space using the Poincaré ball model. 
\begin{dfn}[Poincaré ball model]
We define the following space $\mathcal{H}_p^D$.
\begin{equation}
    \mathcal{H}_p^D\overset{\mathrm{def}}{=}\left\{x\mid x \in \mathbb{R}^{D},x_1^2 +\cdots + x_D^2 <1\right\}.
\end{equation}
We set $x=(x_1,\cdots,x_D)^\top$. The $D$-dimensional Poincaré ball model is a space equipped with the following Riemannian metric defined on $\mathcal{H}_p^D$.
\begin{equation}
    g\overset{\mathrm{def}}{=}4\frac{dx_1^2+\cdots+dx_D^2}{(1-x_1^2-\cdots-x_D^2) ^2}.
\end{equation}
\end{dfn}
The Poincaré ball model is equivalent to other hyperbolic models such as the Lorentz model.
\begin{dfn}[Lorentz model]
\label{Lorentz_M}
We define the following space $\mathcal{H}_l^D$.
\begin{equation}
    J_D \stackrel{\text { def }}{=}\left[\begin{array}{cccc}
-1 & 0 & \ldots & 0 \\
0 & 1 & & \\
\vdots & & \ddots & \\
0 & & & 1
\end{array}\right] ,
\end{equation}
\begin{equation}
    \langle x, y\rangle_{\mathcal{L}} \stackrel{\text { def }}{=} x^{\top} J_D y ,
\end{equation}
\footnotesize
\begin{equation}
    \mathcal{H}^D_l \stackrel{\text { def }}{=}\left\{x=\left(x_0, \cdots, x_D\right)^T \mid x \in \mathbb{R}^{D+1},\langle x, x\rangle_{\mathcal{L}}=-1, x_0>0\right\} .
\end{equation}
\normalsize
The $D$-dimensional Lorentz model is a space equipped with the following Riemannian metric defined on $\mathcal{H}_l^D$.
\begin{equation}
    g\overset{\mathrm{def}}{=}-(dx_0)^2+dx_1^2+\cdots+dx_D^2.
\end{equation}
\end{dfn}
A Riemannian manifold $\mathcal{M}$ is a differentiable manifold equipped with a Riemannian metric $g_\mathcal{M}(x)$, which assigns an inner product to the tangent space at each point $x\in\mathcal{M}$. The notion of distance on a Riemannian manifold is defined via the Riemannian metric, and integration over the manifold is carried out with respect to a volume element induced by this metric. The definition of the volume element on a Riemannian manifold is provided below.
\begin{equation}
\label{volume_form}
d\mathrm{vol}(x) \overset{\mathrm{def}}{=} \sqrt{\det g_{\mathcal{M}}(x)} \, dx.
\end{equation}
\bluetext{Let $x^n$ denote a sequence of data points on a Riemannian manifold. The volume element of the n-fold product manifold is defined as follows. 
\begin{equation}
\label{volume_form_n}
\begin{aligned}
\sqrt{\det g_{\mathcal{M}}(x^n)} \ &dx^n \\
\overset{\mathrm{def}}{=}
&\sqrt{\det g_{\mathcal{M}}(x_1)} \times \cdots  \times \sqrt{\det g_{\mathcal{M}}(x_n)}\ dx^n.
\end{aligned}
\end{equation}
}
The Riemannian metric $g_\mathcal{M}(x)$ depends on the coordinate system defined on the Riemannian manifold. Consequently, the volume element also becomes a coordinate-dependent quantity. The volume elements defined under two different coordinate systems transform according to the product with the Jacobian determinant between the coordinate systems. From this perspective, functions defined with respect to the volume element are coordinate invariant, since the Jacobian arising from a coordinate transformation is absorbed into the transformation of the volume element itself. Henceforth, probability density functions defined with respect to the volume element will be denoted using the subscript “vol”, such as $p_{\mathrm{vol}}$. The probability density function defined with respect to the volume element and that defined with respect to the coordinate differentials transform as follows.
\bluetext{
\begin{equation}
\label{vol_chage_det}
    p(x^n \mid \theta)=p_{\mathrm{vol}}(x^n \mid \theta) \sqrt{\det g_{\mathcal{M}}(x^n)}.
\end{equation}}
As a probability density function defined with respect to the volume element on a Riemannian manifold, the Riemannian Gaussian Distribution (R-GD) has been proposed to generalize the classical normal distribution \cite{said2017gaussian}.
\begin{exa}[R-GD]
\label{R_GD1}
Let $x,\mu\in\mathcal{M}, \sigma\in\mathbb{R}$. Let $d(\cdot,\cdot)$ denote the distance between two points induced by the Riemannian metric. On a Riemannian manifold, the probability density function representing a normal distribution, referred to as the Riemannian Gaussian distribution (R-GD), is defined with respect to the volume element as follows.
\begin{equation}
\label{R_GD}
    p_{\mathrm{vol}}(x \mid \mu, \sigma) \overset{\mathrm{def}}{=} \frac{1}{\xi(\sigma)} \exp\left( -\frac{1}{2\sigma^2} d^2(x, \mu) \right),
\end{equation}
\begin{equation}
    \xi(\sigma) \stackrel{\text { def }}{=} \int_{\mathcal{M}} \exp \left(-\frac{1}{2 \sigma^2} d^2(x, \mu)\right) d\mathrm{vol}(x) .
\end{equation}
Here, let $D$ denote the dimension of the Riemannian manifold $\mathcal{M}$. Then, the parameter space of R-GD has dimension $D+1$.
\end{exa}

\subsection{Conventional NML}
Let $\mathcal{X}$ be the data space, and denote the product set of $n$ copies of $\mathcal{X}$ by $\mathcal{X}^n$. Let $\Theta$ be the parameter space, and $\mathcal{P}=\left\{p( x\mid \theta):x\in\mathcal{X}, \theta\in\Theta\right\}$ be the probability model class. The NML distribution is defined by normalizing the likelihood specified by the maximum likelihood estimator. Let $x^n\in\mathcal{X}^n$ be given. The NML distribution is defined as follows \cite{shtar1987universal}. Here, $\hat\theta(x^n)=\operatorname{argmax}_{\theta\in\Theta} p(x^n \mid \theta)$ denotes the maximum likelihood estimator.
\begin{equation}
\label{NML_distribution}
  p_{\mathrm{NML}}(x^n) = \frac{p(x^n \mid \hat{\theta}(x^n))}{\int_{y^n \in \mathcal{X}^n} p(y^n \mid \hat{\theta}(y^n)) \, dy^n}.
\end{equation}
The NML distribution defined as above constitutes a valid probability distribution over the sample space $\mathcal{X}^n$. The denominator in Equation (\ref{NML_distribution}) is referred to as the parametric complexity (PC), which is generally known to be
computationally intractable. We denote the PC as follows. 
\begin{equation}
\label{PC_label}
    \mathcal{C}(\mathcal{P})\stackrel{\text { def }}{=} \int_{y^n \in \mathcal{X}^n} p(y^n \mid \hat{\theta}(y^n)) \, dy^n.
\end{equation}
The NML code-length is defined as the negative logarithm of the NML distribution. \bluetext{Throughout this paper, all logarithms are taken to base 2.}
\bluetext{
\begin{equation}
\label{NML_codelengh}
\begin{aligned}
  \mathcal{L}_{\mathrm{NML}}(x^n \mid \mathcal{P})&=-\log p_{\mathrm{NML}}(x^n)\\
  &= -\log p(x^n \mid \hat{\theta}(x^n)) + \log \mathcal{C}(\mathcal{P}).
\end{aligned}
\end{equation}}
\bluetext{The NML code-length is established as the SC, and the above expression relates the SC to the PC.} The NML code-length achieves the following minimax regret \cite{shtar1987universal}. Let $q$ be a probability density function defined on $\mathcal{X}^n$.
\begin{equation}
\label{min_max_E}
\begin{aligned}
    &p_{\mathrm{NML}}(\cdot)\\
    &=\underset{q}{\operatorname{argmin}} \max _{x^n}\left\{-\log q(x^n)-\min _\theta(-\log p(x^n \mid \theta))\right\}.
\end{aligned}
\end{equation}

\section{The coordinate dependence of NML}
\label{Sec3}
The definition of conventional NML has been provided above. However, the conventional NML formulation lacks coordinate invariance. When the data sequence $x^n$ is transformed into another sequence $y^n$, the Jacobian determinant arising from the coordinate transformation appears in the formulation, as shown below.
\begin{equation}
\label{NML_jacobian}
p_{\mathrm{NML}}(x^n) \left| \frac{\partial x^n}{\partial y^n} \right| = p_{\mathrm{NML}}(y^n).
\end{equation}
Since most previous studies dealing with the MDL principle have been conducted on Euclidean spaces with natural coordinate systems, the lack of coordinate invariance in the NML code-length has not posed a significant issue. However, in the case of Riemannian manifolds, standard coordinate systems do not generally exist, and probability density functions depend on the choice of coordinates. As a result, interpreting the negative logarithm of a probability density as a code-length leads to coordinate-dependent and thus non-unique values. We present below an example in which the probability density function and the code-length vary under a coordinate transformation.
\begin{exa}
Hyperbolic space can be represented by either the Lorentz model or the Poincaré ball model, which are equivalent representations.
Suppose a probability density function $f_L(x), x\in\mathcal{H}_l^D$ is defined on the Lorentz model. To express this density in the Poincaré ball model by $f_P(x)$, we apply the corresponding coordinate transformation, yielding the following form. 
\begin{equation}
    f_P(x) = f_L(x) \sqrt{1+|x|^2}\left(\frac{2}{1-|x|^2}\right)^D,
\end{equation}
\begin{equation}
    \begin{aligned}
        -\log f_P(x) +&\log f_L(x) \\
        &=-\log \left(\sqrt{1+|x|^2}\left(\frac{2}{1-|x|^2}\right)^D\right).
    \end{aligned}
\end{equation}
\end{exa}
Now that the data space has expanded from Euclidean spaces to Riemannian manifolds, it is necessary for the NML code-length to be treated as a coordinate-invariant quantity. In fact, when applying the NML code-length to tasks such as comparing different Riemannian manifolds or detecting changes under settings not aligned with the natural parameterization of Euclidean space, the coordinate dependence can significantly affect performance. In this study, we propose Rm-NML as an extension of the conventional NML framework, designed to operate on Riemannian manifolds while maintaining coordinate invariance. This formulation enables the application of NML-based model selection and code-length analysis in non-Euclidean settings, where standard coordinate systems may not exist and the geometry of the data space plays a crucial role.

\section{Riemannian manifold NML}
\label{Sec4}
In this study, we propose Rm-NML as an NML that works on Riemannian manifolds with coordinate invariance. Rm-NML is constructed from a probability density function over volume elements, and it is itself a probability density function on volume elements. The definition is as follows.
\begin{dfn}[Riemannian manifold NML]
\label{Rm_NML_def22}
  We consider a Riemannian manifold data space $\mathcal{M}$. Let $x^n\in\mathcal{X}^n$ denote a data sequence defined on the Riemannian manifold $\mathcal{M}$. Let $p_{\mathrm{vol}}(x^n \mid \theta)$ be a probability density function on $\mathcal{M}^n$, defined with respect to the product volume element induced by the Riemannian structure on $\mathcal{M}$. Let $\hat{\theta}$ be the maximum likelihood estimator of $\theta$. The Rm-NML distribution is then defined as follows.
  \begin{equation}
  \label{Rm_NML_def}
p_{\mathrm{Rm\text{-}NML}}(x^n) \overset{\mathrm{def}}{=} 
\frac{p_{\mathrm{vol}}(x^n \mid \hat{\theta}(x^n))}{
\int_{y^n \in \mathcal{X}^n} p_{\mathrm{vol}}(y^n \mid \hat{\theta}(y^n)) \, d\mathrm{vol}(y^n)}.
  \end{equation}
\end{dfn}
By the transformation in Equation (\ref{volume_form}), it can be seen that the denominator of the above Rm-NML distribution coincides with the PC defined in Equation (\ref{PC_label}).
\begin{equation}
\label{PC_label2}
    \mathcal{C}(\mathcal{P})= \int_{y^n \in \mathcal{X}^n} p_{\mathrm{vol}}(y^n \mid \hat{\theta}(y^n)) \, d\mathrm{vol}(y^n).
\end{equation}
The negative logarithm of the Rm-NML is defined as the Rm-NML code-length. 
\begin{equation}
\label{RMNML_CL}
    \mathcal{L}_{\mathrm{Rm-NML}}\left(x^n\right) \overset{\mathrm{def}}{=} -\log p_{\mathrm{Rm\text{-}NML}}(x^n).
\end{equation}

The Rm-NML also constitutes a valid probability density function with respect to the volume element on 
$\mathcal{X}^n$. Therefore, the Rm-NML code-length can be interpreted as the information content of a probability density function defined with the volume element as its underlying measure.
The Rm-NML achieves the following minimax regret.
\begin{equation}
\label{min_max_RmNML}
\begin{aligned}
    &p_{\mathrm{Rm\text{-}NML}}(\cdot)\\
    &=\underset{q_{\mathrm{vol}}}{\operatorname{argmin}} \max _{x^n}\left\{-\log q_{\mathrm{vol}}(x^n)-\min _\theta(-\log p_{\mathrm{vol}}(x^n \mid \theta))\right\}.
\end{aligned}
\end{equation}
The worst-case regrets in Equations (\ref{min_max_RmNML}) and (\ref{min_max_E}) are equivalent, as indicated by the following relationship.
\begin{equation}
\label{max_reg_chagne}
\begin{aligned}
    &\max _{x^n}\left\{-\log q(x^n)-\min _\theta(-\log p(x^n \mid \theta))\right\}\\
    &=\max _{x^n}\left\{-\log q_{\mathrm{vol}}(x^n)-\min _\theta(-\log p_{\mathrm{vol}}(x^n \mid \theta))\right\}.
\end{aligned}
\end{equation}
The denominator of the Rm-NML coincides with that of the conventional NML, and thus corresponds to the PC. This observation implies that Rm-NML achieves the same worst-case regret as the conventional NML.

\subsection{Justification of Rm-NML as a Code-length}
The Rm-NML distribution defined in this study is itself a valid probability density function with respect to the Riemannian volume element. By taking its negative logarithm, we obtain the Rm-NML code-length. In this section, we provide a theoretical justification for interpreting the negative log-density—defined with respect to the Riemannian volume element—as a valid code-length.

Consider a continuous space $\mathcal{X}$ and an element $x\in\mathcal{X}$. Consider $\mu$ as a measure. Let $p(x)$ be a probability density function defined on $\mathcal{X}$. Suppose that $\rho= \{S_1, S_2, \cdots \}$ is a countable and mutually exclusive partition of $\mathcal{X}$. Under these assumptions, the following two theorems hold.
\begin{thm}[Existence]
\label{IFJ1}
    Consider a countable and mutually exclusive partition $\rho$ of the set $\mathcal{X}$. There exists a prefix code for $p(x)$ such that the length $\ell_S$ of the codeword assigned to each region $S \in \partition$ satisfies the following condition. Here, let $\vol(S)$ denote the volume of the region S.
    \begin{equation}
    \ell_S = \ceil{\supS{-\log_2 \pdf(x)} - \log_2 \vol(S)}.
    \end{equation}
\end{thm}

\begin{thm}[Optimality in terms of expected code-length]
\label{IFJ2}
    The average code-length $\bar{L}$ of any instantaneous uniquely decodable code with respect to the partition $\rho$ is lower bounded by $L_{\text{lower}}$, which is defined as follows. Here, let $\vol(S)$ denote the volume of the region S.
    \begin{equation}
    \begin{aligned}
        L_{lower} &= \E_S \left[ \infS{-\log_2 \pdf(x)} - \log_2 \vol(S) \right] \\
        &= \sum_{S \in \partition} P(S) \left( \infS{-\log_2 \pdf(x)} - \log_2 \vol(S) \right). 
    \end{aligned}
    \end{equation}
\end{thm}
When the above two theorems hold and each element of the partition is sufficiently small, both $\sup_{x \in S} \pdf(x) \approx \pdf(x_S)$ and $\inf_{x \in S} \pdf(x) \approx \pdf(x_S)$ can be approximated by the value of the density at a representative point $x_S \in S$. Under this approximation, the proposed code-length $\ell_S$ can be approximated as follows, and it possesses both existence and optimality in terms of the average code-length.
\begin{equation}
\begin{aligned}
    \ell_S &\approx \lceil -\log_2 \pdf(x_S) - \log_2 \vol(S) \rceil \\&= \lceil -\log_2 (\pdf(x_S) \vol(S)) \rceil.
\end{aligned}
\end{equation}
The definition of information content for a continuous probability density function $-\log p(x)\Delta x = -\log p(x)-\log \Delta x$ is typically derived by considering the information content of the probability $p(x)\Delta x$ over an infinitesimal region $\Delta x$, and neglecting the term associated with $\Delta x$ itself, thereby isolating the component that depends solely on $p(x)$. When considering a partition with sufficiently small elements, the volume $\vol(S)$ becomes infinitesimal. Following the standard definition of information content for continuous probability density functions, this infinitesimal value is ignored in the present study. As a result, a code-length that depends solely on the probability density function, and not on the volume element, can be constructed.

\subsection{Relationship with previous studies}
As discussed earlier, when applying the MDL principle to data spaces modeled as Riemannian manifolds, it is essential that the resulting code-length possesses coordinate invariance. However, prior studies and previous applications of the MDL principle have largely overlooked the coordinate dependence of the NML formulation. The Rm-NML defined in this study does not refute the existing body of research based on the conventional NML, which lacks coordinate invariance. Rather, by introducing Rm-NML as a coordinate-invariant extension, this work clarifies that the coordinate dependence of conventional NML arises solely from the underlying volume element.
\begin{equation}
\label{NML_RmNML}
\begin{aligned}
    - \log p_{\mathrm{NML}}(x^n) -& \left( - \log p_{\mathrm{Rm\text{-}NML}}(x^n) \right) \\
    &= - \log \sqrt{\det g_{\mathcal{M}}(x^n)}.
\end{aligned}
\end{equation}
Most of the previous studies based on the MDL principle have been conducted under the standard setting of Euclidean space. In this setting, the volume element is identically 1 at every point, and thus the \bluetext{Rm-NML coincides} with the conventional NML. In other words, the Rm-NML generalizes the standard NML formulation. Furthermore, even when a coordinate system other than the standard Cartesian coordinate system—such as polar coordinates—is employed on Euclidean space, the value of the volume element remains unchanged as long as the data points are fixed. Consequently, the coordinate dependence does not affect model selection performance in such cases.

\section{Calculation of PC on Riemannian manifolds}
\label{Sec5}
The denominator of the Rm-NML is the PC, and its value is derived equivalently from the definition of PC in Euclidean space. In general, PC is known to be computationally intractable, and various techniques have been developed to approximate or compute it. However, once the data space is extended to a Riemannian manifold, it is non-trivial whether existing computational methods remain applicable. In particular, properties that hold only in Euclidean space are no longer valid in this setting. In this study, we extend existing methods for computing PC so that they are applicable on Riemannian data spaces. We verify the validity of these methods under this generalization. Moreover, since the extension to Riemannian manifolds is expected to make the computation of PC even more challenging, we also propose multiple alternative approaches that facilitate the derivation of its value.
\subsection{Asymptotic Approximation Method}
One method for computing the PC is to use an asymptotic approximation, which becomes effective when the number of data points $n$ is sufficiently large \cite{rissanen2002fisher}. The method for computing the PC using the asymptotic approximation proposed in \cite{rissanen2002fisher} holds under the following five conditions. 

\begin{thm}[Asymptotic approximation formula for PC \cite{rissanen2002fisher}]
\label{PC_Asy}
First, the Fisher information matrix is defined as follows.

\begin{equation}
\label{fisher_E}
    I(\theta) \overset{\mathrm{def}}{=} \lim_{n \to \infty} \frac{1}{n} E^n_{\theta} \left[- 
     \left(\frac{\partial^2 \log p(x^n\mid \theta)}{\partial \theta \partial \theta^\top}\right)
\right].
\end{equation}

\begin{enumerate}
\item The parameter space $\Theta$ is assumed to be compact. There exist positive constants $c_1$ and $c_2$ such that for all $\theta\in\Theta$, the inequality $0<c_1\le\lvert I(\theta) \rvert\le c_2<\infty$ holds, where $\lvert I(\theta) \rvert$ denotes the determinant of the Fisher information matrix.
\item The Fisher information matrix $I(\theta)$ is continuous with respect to $\theta$.
\item $\int \sqrt{|I(\theta)|} \, d\theta < \infty$.
\item The asymptotic normality of the maximum likelihood estimator holds uniformly over all points $\theta\in\Theta$. Let $\hat{\theta}(x^n)$ denote the maximum likelihood estimator of $\theta$ based on the data $x^n$.
\begin{equation}
\sqrt{n} \left( \hat{\theta}(x^n) - \theta \right) \xrightarrow[n \to \infty]{\; \; \; \; \; \; \; \; } \mathcal{N} \left( 0, I^{-1}(\theta) \right).
\end{equation}
\item There exists a finite positive definite matrix $C_0$ such that, for all $n$, and for any $x^n$ for which the maximum likelihood estimator $\hat{\theta}(x^n)$ belongs to $\Theta$, the following inequality holds.
\begin{equation}
    \frac{1}{n} \left[ 
    - \frac{\partial^2 \log p(x^n\mid \theta)}{\partial \theta \partial \theta^\top}\right]< C_0.
\end{equation}
\end{enumerate}
Under these five conditions, the PC ($\mathcal{C}(\mathcal{P})$) defined in Equation (\ref{PC_label}) is computed as follows. Here, let $k$ denote the dimension of the parameter space $\Theta$.
\begin{equation}
    \log \mathcal{C}(\mathcal{P}) = \frac{k}{2} \log \frac{n}{2\pi} + \log \int_{\theta\in\Theta} \sqrt{|I(\theta)|} \, d\theta + o(1).
\end{equation}
Here, $o(1)$ is taken to be $\lim_{n\to\infty}o(1)=0$.
\end{thm}
\bluetext{
\begin{remark}
In this work, we follow Rissanen's (1996) \cite{rissanen2002fisher} definition of the Fisher information matrix in terms of the second-order derivatives of the log-likelihood, as given in Equation~(\ref{fisher_E}). Alternatively, the Fisher information can also be defined via the first-order derivatives (the score function):
\begin{equation}
I(\theta)
=
\mathbb{E}_{\theta}
\!\left[
\left(
\frac{\partial}{\partial\theta}
\log p(x\mid\theta)
\right)
\left(
\frac{\partial}{\partial\theta}
\log p(x\mid\theta)
\right)^{\mathsf T}
\right].
\end{equation}
Let $(\Omega,\mathcal{F},\mathbb{P})$ be a probability space and let $X:\Omega\to\mathcal{X}$ be a random variable with density $f(x;\theta)$. These two definitions coincide under the following conditions:
\begin{enumerate}
    \item The data sequence $x_1, \dots, x_n$ is independent and identically distributed (i.i.d.).
    \item For the parameter space $\Theta$, the density $f(x;\theta)$ is of class $C^2$ with respect to $\theta$:
    \[
        \theta \mapsto f(X(\omega);\theta) \in C^2, \quad \forall \omega \in \Omega \text{ a.s.}
    \]
    \item For each $\theta \in \Theta$ and $\mu$-almost every $x \in \mathcal{X}$, we have $f(x;\theta) > 0$.
    \item The following integrability conditions hold:
    \[
        \int_{\mathcal{X}} \sup_{\tilde{\theta} \in \mathcal{N}} \bigl\| \nabla_\theta f(x;\tilde{\theta}) \bigr\| \, \mu(dx) < \infty, 
    \]
    \[
        \int_{\mathcal{X}} \sup_{\tilde{\theta} \in \mathcal{N}} \bigl\| \nabla^2_\theta f(x;\tilde{\theta}) \bigr\| \, \mu(dx) < \infty,
    \]
    where $\mathcal{N}$ is a neighborhood of $\theta$.
\end{enumerate}
When these conditions are satisfied, the log-likelihood for i.i.d.\ data decomposes as
\begin{equation}
\log p(x^n\mid\theta)
=
\sum_{i=1}^n \log p(x_i\mid\theta),
\end{equation}
and under the framework of Rissanen, the Fisher information matrix based on second-order derivatives coincides with that based on first-order derivatives in the limit of large $n$:
\begin{equation}
\begin{aligned}
 \lim_{n \to \infty} \frac{1}{n} E^n_{\theta} &\left[- 
     \frac{\partial^2 \log p(x^n\mid \theta)}{\partial \theta \partial \theta^\top}
\right]\\
&=
\mathbb{E}_{\theta}
\!\left[
\left(
\frac{\partial}{\partial\theta}
\log p(x\mid\theta)
\right)
\left(
\frac{\partial}{\partial\theta}
\log p(x\mid\theta)
\right)^{\mathsf T}
\right].
\end{aligned}
\end{equation}
\end{remark}
}

\begin{comment}
\bluetext{\begin{remark}
The Fisher information matrix is defined, following Rissanen (1996)\cite{rissanen2002fisher}, in terms of the second-order derivatives of the log-likelihood as Equation (\ref{fisher_E}). When the data sequence $x_1,\dots,x_n$ is i.i.d., the log-likelihood decomposes as
\begin{equation}
\log p(x^n\mid\theta)
=
\sum_{i=1}^n \log p(x_i\mid\theta).
\end{equation}
In this case, the Fisher information matrix coincides with the definition based on the first-order derivatives (i.e., the score function),
\begin{equation}
I(\theta)
=
\mathbb{E}_{\theta}
\!\left[
\left(
\frac{\partial}{\partial\theta}
\log p(x\mid\theta)
\right)
\left(
\frac{\partial}{\partial\theta}
\log p(x\mid\theta)
\right)^{\mathsf T}
\right].
\end{equation}
Under standard regularity conditions, these two definitions are equivalent in the i.i.d.\ setting, namely,
\begin{equation}
\begin{aligned}
 \lim_{n \to \infty} \frac{1}{n} E^n_{\theta} &\left[- 
     \left(\frac{\partial^2 \log p(x^n\mid \theta)}{\partial \theta \partial \theta^\top}\right)
\right]
\\&=
\mathbb{E}_{\theta}
\!\left[
\left(
\frac{\partial}{\partial\theta}
\log p(x\mid\theta)
\right)
\left(
\frac{\partial}{\partial\theta}
\log p(x\mid\theta)
\right)^{\mathsf T}
\right].
\end{aligned}
\end{equation}
\end{remark}}
\end{comment}
\bluetext{By Theorem \ref{PC_Asy} and Equation (\ref{NML_codelengh}), the NML code-length on a Euclidean data space is given as follows.
\begin{cor}[The NML Code-Length on Euclidean Manifold Data Spaces]
Under the same conditions as in Theorem \ref{PC_Asy}, the NML code-length on a Euclidean data space can be approximated as follows.
\begin{equation}
\label{NML_codelengh_2}
\begin{aligned}
  \mathcal{L}_{\mathrm{NML}}(x^n \mid \mathcal{P})=&-\log p(x^n \mid \hat{\theta}(x^n))\\
  &+\frac{k}{2} \log \frac{n}{2\pi} + \log \int_{\theta\in\Theta} \sqrt{|I(\theta)|} \, d\theta + o(1).
\end{aligned}
\end{equation}
\end{cor}}

However, the PC estimation method based on the above asymptotic approximation cannot be directly extended to Riemannian manifolds due to limitations in both its assumptions and formulation.

\begin{itemize}
\item First, the asymptotic approximation formula for the PC presented in \cite{shtar1987universal} considers a probability density function defined with respect to a density defined by the coordinate system in the Euclidean setting. In contrast, this study extends the data space to a Riemannian manifold and, accordingly, considers probability density functions defined with respect to the volume element induced by the Riemannian metric.
\item By considering probability density functions defined with respect to the volume element, it becomes necessary to carefully define the Fisher information matrix.
\item It is straightforward to verify that the asymptotic approximation of the PC provided in \cite{shtar1987universal} is coordinate-invariant. However, it remains unproven whether the conditions required for the validity of the above theorem \ref{PC_Asy} are preserved under continuous coordinate transformations.
\item The derivation of the asymptotic approximation relies on quantization of the parameter space using hyperrectangles. Such an approach does not hold in curved spaces, such as Riemannian manifolds, where the notion of hyperrectangles is not directly applicable.
\end{itemize}
Specifically, we first demonstrate that the Fisher information matrix defined with respect to a probability density function over a density defined by the coordinate system is equivalent to that defined with respect to a probability density function over volume elements. Next, we explicitly specify the coordinate systems of both the data space and the parameter space on the manifold, and prove that the resulting expressions are invariant under smooth coordinate transformations. By making the coordinate systems explicit, the analysis based on hyperrectangles becomes applicable. Furthermore, we show that the assumptions required for the approximation remain valid under this setting. By establishing these two key points, we derive an asymptotic approximation method for the PC on Riemannian manifolds that is independent of the choice of coordinates. In this study, we do not restrict the parameter space to be Euclidean. The proposed formulation is given as follows.

\begin{thm}[Asymptotic approximation formula for PC on Riemannian manifolds]
\label{Asymptotic_Rm}
Let $p_\mathrm{vol}$ be a probability density function defined on the volume element of the Riemannian manifold $\mathcal{M}$. Let $x\in\mathcal{X}\subset\mathcal{M}$ and $\theta\in\Theta$. Set a coordinate system $\phi$ on $\Theta$ to represent $\theta$. Let the Fisher information matrix with respect to coordinate system $\phi$ be $I_{\phi}(\theta)$.

\begin{equation}
    I_\phi\left(x^n, \theta\right)=-\frac{1}{n}\left[\frac{\partial^2 \log p_{\mathrm{vol}}\left(x^n \mid \phi(\theta)\right)}{\partial \phi(\theta) \partial \phi(\theta)^{\top}}\right],
    \end{equation}
    \begin{equation}
    I_\phi(\theta) \stackrel{\text { def }}{=} \lim _{n \rightarrow \infty} E_\theta^n\left[I_\phi\left(x^n, \theta\right)\right].
\end{equation}
\begin{enumerate}
\item The parameter space $\Theta$ is assumed to be compact. There exist positive constants $c_1$ and $c_2$ such that for all $\theta\in\Theta$, the inequality $0<c_1\le\lvert I_\phi(\theta) \rvert\le c_2<\infty$ holds, where $\lvert I_\phi(\theta) \rvert$ denotes the determinant of the Fisher information matrix.
\item The Fisher information matrix $I_\phi(\theta)$ is continuous with respect to $\theta$.
\item $\int _{\theta\in\Theta}\sqrt{|I_\phi(\theta)|} \, d\theta < \infty$.
\item The asymptotic normality of the maximum likelihood estimator holds uniformly over all points $\theta\in\Theta$. Let $\hat{\theta}(x^n)$ denote the maximum likelihood estimator of $\theta$ based on the data $x^n$.
\begin{equation}
\sqrt{n} \left( \phi(\hat{\theta}(x^n)) - \phi(\theta) \right) \xrightarrow[n \to \infty]{\; \; \; \; \; \; \; \; } \mathcal{N} \left( 0, I^{-1}_\phi(\theta) \right).
\end{equation}
\item There exists a finite positive definite matrix $C_0$ such that, for all $n$ , and for any $x^n$ for which the maximum likelihood estimator $\hat{\theta}(x^n)$ belongs to $\Theta$, the following inequality holds.
\begin{equation}
   I_\phi\left(x^n, \theta\right)< C_0.
\end{equation}
In this case, the asymptotic approximation formula for PC ($\mathcal{C}(\mathcal{P})$) in Equation (\ref{PC_label2}) can be calculated as follows.
\begin{equation}
\label{PC_R}
    \log \mathcal{C}(\mathcal{P}) = \frac{k}{2} \log \frac{n}{2\pi} + \log \int_{\theta\in\Theta} \sqrt{|I_\phi(\theta)|} \, d\phi(\theta) + o(1).
\end{equation}
\end{enumerate}
See Appendix \ref{Ap1} for the proof. 
\end{thm}

\bluetext{By Theorem \ref{Asymptotic_Rm} and Equation (\ref{RMNML_CL}), the Rm-NML code-length on a Riemannian manifold data space is given as follows.
\begin{cor}[The Rm-NML Code-Length on Riemannian Data Spaces]
Under the same conditions as in Theorem \ref{Asymptotic_Rm}, the Rm-NML code-length on a Riemannian Manifold data space can be approximated as follows.
\begin{equation}
\label{NML_codelengh_3}
\begin{aligned}
  \mathcal{L}_{\mathrm{Rm-NML}}&(x^n \mid \mathcal{P})=-\log p_{\mathrm{vol}}(x^n \mid \hat{\theta}(x^n))\\
  &+\frac{k}{2} \log \frac{n}{2\pi} + \log \int_{\theta\in\Theta} \sqrt{|I_\phi(\theta)|} \, d\phi(\theta) + o(1).
\end{aligned}
\end{equation}
\end{cor}}

This expression makes explicit the coordinate system of the parameter $\theta$ in the existing asymptotic approximation. Based on the nontrivial aspects encountered in extending the existing asymptotic approximation method to a Riemannian manifold, we present the derivation of the asymptotic approximation formula for PC on a Riemannian manifold in the form given above. Equation (\ref{vol_chage_det}) shows that the difference between the probability density functions defined with respect to infinitesimal quantities and those defined with respect to the volume element lies in the volume element itself. Since the volume element is independent of the parameter $\theta$, the Fisher information matrices computed from these two probability density functions are identical.

In conventional asymptotic approximation methods, the parameter space was quantized using hyperrectangles. By introducing a coordinate system $\phi$ on the parameter space $\Theta$, this procedure can be translated into a discussion on Euclidean space.
\begin{equation}
    \phi(\theta) \in \phi(\Theta) \subset \mathbb{R}^k.
\end{equation}
Based on the above three points, the existing asymptotic approximation can be extended to a Riemannian manifold data space in the form of Theorem\ref{Asymptotic_Rm}.

Theorem \ref{Asymptotic_Rm} enables the asymptotic approximation of the value of the PC on a Riemannian manifold in a coordinate-explicit form. However, the Rm-NML distribution defined in this study possesses coordinate invariance, and therefore the computation of the PC, which constitutes the denominator of the distribution, must also be coordinate-invariant. Furthermore, the five regularity conditions assumed in the derivation of the asymptotic approximation must remain valid under continuous coordinate transformations. By demonstrating these two aspects of coordinate invariance, the Rm-NML code-length can be shown to possess coordinate invariance as a model selection criterion.

We first verify the coordinate invariance of the computed result. 
\bluetext{\begin{proposition}[The asymptotic approximation of the PC on a Riemannian manifold is invariant under coordinate transformations.]
\label{fisher_change1}
    Let $\phi$ and $\psi$ be coordinate systems related by a continuously differentiable coordinate transformation. The asymptotic approximation of the PC on a Riemannian manifold yields consistent results across these coordinate systems. In particular, the following holds for the second term on the right-hand side of Equation (\ref{PC_R}).  
\begin{equation}
\log \int \sqrt{\left|I_\phi(\theta)\right|} d \phi(\theta)  =\log \int \sqrt{\left|I_{\psi}(\theta)\right|} d \psi(\theta).
\end{equation}
\begin{proof}
Let $\phi$ and $\psi$ be coordinate systems related by a continuously differentiable coordinate transformation. Under such a transformation, the determinant of the Fisher information matrix transforms as follows.
\begin{equation}
\label{fisher_change}
    \left|I_\phi(\theta)\right|=\left|\frac{\partial \psi}{\partial \phi}\right|^2\left|I_{\psi}(\theta)\right| .
\end{equation}
Using Equation (\ref{fisher_change}), we apply the coordinate transformation to the second term on the right-hand side of Equation (\ref{PC_R}).
\begin{equation}
    \begin{aligned}
\log \int \sqrt{\left|I_\phi(\theta)\right|} d \phi(\theta) & =\log \int \sqrt{\left|I_{\psi}(\theta)\right|}\left|\frac{\partial \psi}{\partial \phi}\right| d \phi(\theta) \\
& =\log \int \sqrt{\left|I_{\psi}(\theta)\right|} d \psi(\theta).
\end{aligned}
\end{equation}
\end{proof}
\end{proposition}}
\bluetext{
\begin{remark}
    The statement of Proposition \ref{fisher_change1} is obvious if we regard the determinant $|I_\phi (\theta)|$ of the Fisher metric as a volume form. Nevertheless, we have provided the explicit form and a coordinate-based proof for readers' convenience.
\end{remark}}
From the result of the above coordinate transformation, it follows that the asymptotic approximation of the PC derived in this study is invariant with respect to the choice of coordinates. In this study, we also prove that the five regularity conditions in Theorem \ref{PC_Asy} are invariant under coordinate transformations. 

\begin{thm}[Coordinate invariance of the regularity conditions]
    Let $\phi$ and $\psi$ be two coordinate systems related by a twice continuously differentiable coordinate transformation. If the five regularity conditions in Theorem \ref{PC_Asy} hold in the coordinate system $\phi$, then they also hold in the coordinate system $\psi$. See Appendix \ref{Ap2} for the proof.
\end{thm}

In this study, we confirmed that if Conditions 1 through 5 hold in a coordinate system $\phi$ , they also hold in a coordinate system $\psi$ related to $\phi$ via a twice continuously differentiable transformation. Condition 2 is trivial, and the invariance of Condition 3 directly follows from the coordinate invariance of the associated quantity. For Conditions 1 and 5, it can be proven that they are preserved under coordinate transformations based on the transformation of the Fisher information matrix given in Equation (\ref{fisher_change}) and the compactness of the parameter space $\Theta$. Condition 4 is often nontrivial. In this study, we formulated the uniform validity of the asymptotic normality of the maximum likelihood estimator in the form given below, and proved that this condition is preserved under any twice continuously differentiable coordinate transformation\footnote{Rissanen (1996), who derived the asymptotic approximation of the PC, did not provide a rigorous definition of the 'uniform validity of the asymptotic normality of the maximum likelihood estimator' \cite{rissanen2002fisher}. In this study, we adopt this definition to demonstrate the coordinate invariance of Condition 4; nevertheless, the proof of the asymptotic approximation of the PC given in Rissanen (1996) remains valid under this definition.}. Let $\mathcal{X}^D$ be a measurable subset of $\mathbb{R}^D$ whose boundary has measure zero.
\begin{dfn}[Uniform validity of the asymptotic normality of the maximum likelihood estimator]
\label{Condition4}
    \begin{equation}
        \begin{aligned}
& X \sim \mathcal{N}\left(0, I_\phi(\theta)^{-1}\right), \forall \varepsilon>0, \exists N \in \mathbb{N}, \forall n \in \mathbb{N}, \forall \theta \in \Theta, \\
&\forall A_\phi \in\left\{\mathcal{X}^D\right\} ; n \geq N \\
&\Rightarrow\left|p\left(\sqrt{n}\left(\phi\left(\hat{\theta}\left(x^n\right)\right)-\phi(\theta)\right) \in A_\phi\right)-p\left(X \in A_\phi\right)\right|<\varepsilon .
\end{aligned}
    \end{equation}
\end{dfn}
Based on the definition of the uniform validity of the asymptotic normality of the maximum likelihood estimator given in Definition \ref{Condition4}, the argument presented by Rissanen (1996), who proposed the asymptotic approximation of the PC, is valid \cite{rissanen2002fisher}. Similarly, the derivation of the asymptotic approximation of the PC on a Riemannian manifold proposed in this study is also justified. The formalization of the uniform validity in Definition \ref{Condition4} expresses 'uniform validity' explicitly in terms of uniform convergence.

From the above, Theorem \ref{Asymptotic_Rm} possesses coordinate invariance in both its computational result and its regularity conditions. Consequently, the asymptotic approximation given by Theorem \ref{Asymptotic_Rm} enables the calculation of the PC of a probabilistic model as a value intrinsic to the adopted Riemannian manifold data space, unaffected by the choice of coordinate system.
\subsection{A simple method for computing asymptotic approximations of PC}
An asymptotic approximation of the PC on a Riemannian manifold has been obtained. However, the second term on the right-hand side of Equation (\ref{PC_R}) remains nontrivial and difficult to compute. Among the Riemannian data spaces of interest, spaces with constant curvature—such as hyperbolic spaces and hyperspheres—are of particular relevance. These spaces are instances of Riemannian symmetric spaces. Riemannian symmetric spaces are also Riemannian homogeneous spaces, where isometric transformations act transitively. In this section, we compute the PC for probability density functions defined based on the distance structure of Riemannian homogeneous spaces and Riemannian symmetric spaces. The class of probability density functions considered in this chapter is given as follows.

\bluetext{\begin{dfn}[Distance-determined-type probability distribution family]Let $\mathcal{M}$ be a Riemannian manifold and $d$ be the distance function equipped with $\mathcal{M}$. Also, let $\Gamma$ be a subset of $\mathbb{R}^m$, where $m$ is a nonnegative integer.
Consider a parametric Riemannian probability density function $p_\mathrm{vol}$ on the manifold $\mathcal{M}$ as data space, where $p_\mathrm{vol}$ is parametrized by a parameter in the data space $\mathcal{M}$ and another parameter in $\Gamma$.
In other words, for any $\theta \in \mathcal{M}$ and $\gamma \in \mathbb{R}^m$, $p_\mathrm{vol} (\cdot \mid \theta, \gamma): \mathcal{M} \to \mathbb{R}{\ge 0}$ is a Riemannian probability density function on $\mathcal{M}$.
Remark that $\theta$ is in the data space $\mathcal{M}$, while $\gamma$ may not. We say that $p_\mathrm{vol}$ (or the distribution family represented by $p_\mathrm{vol}$) is a \textit{distance-determined-type probability distribution family} if there exists a function $f: \mathbb{R}{\ge 0} \times \mathbb{R}^m \to \mathbb{R}{\ge 0}$ such that
\begin{equation}
p_{\mathrm{vol}}(x \mid \theta, \gamma)=f\left(d(x, \theta)^2, \gamma\right),
\end{equation}
holds for any $x, \theta \in \mathcal{M}$ and $\gamma \in \Gamma \subset \mathbb{R}^m$.
\end{dfn}}

The squared distance is employed in order to circumvent non-differentiable points such as the origin.

R-GD based on Example \ref{R_GD1} belongs to this family of distributions. When the parameter $\theta$ represents a representative point of the data, such as the centroid, both $\theta$ and a data point $x$ lie in the same space.

\bluetext{\subsubsection{Motivation for distance-determined-type Distribution Families}
We extend the discussion from Euclidean space to Riemannian manifolds precisely because of our interest in their distance structures; accordingly, the probability density functions should depend on the underlying distance. In this sense, distance-determined-type probability distributions constitute a natural class of distribution families on Riemannian manifolds.
If a probability density function does not explicitly depend on distance, then the distance structure of the space may not be fully exploited. This point is particularly important in machine learning applications—most notably in settings that employ hyperbolic space as the data space \cite{nickel2017poincare}. Since hyperbolic space is diffeomorphic to Euclidean space, there is little justification for treating the data space as hyperbolic when the probability density function is independent of distance.}

\subsubsection{Examples of distance-determined-type Distribution Families}
\bluetext{R-GD belongs to the family of distance-determined-type distributions and can reproduce the Gaussian distribution on an arbitrary Riemannian manifold \cite{said2017gaussian}. In Euclidean space, examples of distance-determined-type distribution families include the Gaussian, Student’s t, Laplace, and Cauchy distributions. On the sphere, the von Mises distribution, which models angular or phase-valued data, is an example of a distance-determined-type distribution family \cite{mardia2009directional}. In hyperbolic space, the hyperbolic Student’s t distribution and the hyperbolic Cauchy distribution are commonly used as examples of distance-determined-type distribution families \cite{guo2022co}. In this study, we formulate a theorem that provides a simple method for computing the PC of distance-determined-type distribution families, and this theorem is applicable to all of the distributions mentioned above.}

We first define the relevant Fisher information matrices. Let $\phi(\theta)$ be the coordinate system representing the parameter $\theta$ on the Riemannian symmetric space $\mathcal{M}$. We define the combined parameter vector as $\eta=\left(\phi(\theta)^{\top}, \gamma^{\top}\right)^{\top}$. The vector $\eta$ is treated as a $D+m$ dimensional vector. The Fisher information matrix of the probability density function $p_{\mathrm{vol}}(x \mid \theta, \gamma)$ is denoted by 
$I(\phi(\theta),\gamma)$, and is defined as follows.
\begin{equation}
    I\left(x^n, \phi(\theta),\gamma\right)=-\frac{1}{n}\left[\frac{\partial^2 \log p_{\mathrm{vol}}\left(x^n \mid \eta\right)}{\partial \eta \partial \eta^{\top}}\right] ,
    \end{equation}
    \begin{equation}
    I(\phi(\theta),\gamma) \stackrel{\text { def }}{=} \lim _{n \rightarrow \infty} E_\theta^n\left[I\left(x^n, \phi(\theta),\gamma\right)\right].
\end{equation}
The Fisher information matrix with $\gamma$ fixed is defined as follows.
\begin{equation}
    I_{\phi}\left(x^n, \theta,\gamma\right)=-\frac{1}{n}\left[\frac{\partial^2 \log p_{\mathrm{vol}}\left(x^n \mid \theta,\gamma\right)}{\partial \phi(\theta) \partial \phi(\theta)^{\top}}\right] ,
    \end{equation}
    \begin{equation}
    I_{\phi}(\theta,\gamma) \stackrel{\text { def }}{=} \lim _{n \rightarrow \infty} E_\theta^n\left[I_{\phi}\left(x^n, \theta,\gamma\right)\right].
\end{equation}
The Fisher information matrix with $\theta$ fixed is defined as follows.
\begin{equation}
    I_{\gamma}\left(x^n, \theta,\gamma\right)=-\frac{1}{n}\left[\frac{\partial^2 \log p_{\mathrm{vol}}\left(x^n \mid \theta,\gamma\right)}{\partial \gamma \partial \gamma^{\top}}\right] ,
    \end{equation}
    \begin{equation}
    I_{\gamma}(\theta,\gamma) \stackrel{\text { def }}{=} \lim _{n \rightarrow \infty} E_\theta^n\left[I_{\gamma}\left(x^n, \theta,\gamma\right)\right].
\end{equation}
In this study, we derive a theorem that simplifies the computation of Equation (\ref{PC_R}) using the definitions of the Fisher information matrices given above. We define a normal orthonormal basis at a point $\nu$ on a Riemannian manifold $\mathcal{M}$ as a coordinate system in which the Riemannian metric at $\nu$ is represented by the identity matrix.

\begin{thm}[Asymptotic approximation of probability density functions defined by distance and additional parameters on riemannian symmetric spaces]
\label{Asymptotic_PC_Easy}
We consider a \bluetext{distance-determined-type} probability distribution family defined on a Riemannian manifold $\mathcal{M}$. We define the combined parameter vector as $\eta=\left(\phi(\theta)^{\top}, \gamma^{\top}\right)^{\top}$. Let $\hat{\eta}=\left(\phi(\hat{\theta})^{\top}, \hat{\gamma}^{\top}\right)^{\top}$ denote the maximum likelihood estimator of the parameter. We assume that the five conditions in Theorem \ref{Asymptotic_Rm} also hold with respect to the parameter $\eta$. Based on the above setting, the following results can be established, leading to the derivation of the PC on a Riemannian symmetric space.

\begin{enumerate}

\item For any reference point $\nu\in\mathcal{M}$, we define a coordinate system $\zeta_{\nu}$ on $\nu$ with a normal orthonormal basis. In this setting, fixing $\gamma$ and choosing $\zeta_{\nu}$ as the coordinate system for $\theta$, the determinant of the Fisher information matrix $I_{\zeta_{\nu}}(\theta, \gamma)$ evaluated at $\theta=\nu$, denoted by $|I_{\zeta_{\nu}}(\nu, \gamma)|$, is independent of the choice of $\nu$. We denote this value by $C_{\theta}(\gamma)$.
\begin{equation}
    |I_{\zeta_{\nu}}(\nu, \gamma)|=C_{\theta}(\gamma).
\end{equation}

\item Moreover, the determinant $|I_{\gamma}(\theta, \gamma)|$ is independent of $\theta$. We denote this value by $C_{\gamma}(\gamma)$.
\begin{equation}
    |I_{\gamma}(\theta, \gamma)| = C_{\gamma}(\gamma).
\end{equation}

\item In this case, the asymptotic approximation formula for PC ($\mathcal{C}(\mathcal{P})$) in Equation (\ref{PC_label2}) can be calculated as follows.
\begin{equation}
\begin{aligned}
    \log \mathcal{C}(\mathcal{P}) =& \frac{D+m}{2} \log \frac{n}{2\pi} + \log \mathrm{vol}(\Theta) \\
    &+ \log \int_{\gamma\in\Gamma} \sqrt{C_{\theta}(\gamma)C_{\gamma}(\gamma)}d\gamma + o(1).
\end{aligned}
\end{equation}
\end{enumerate}
See Appendix \ref{Ap3} for the proof.
\end{thm}

\bluetext{By Theorem \ref{Asymptotic_PC_Easy} and Equation (\ref{RMNML_CL}), the Rm-NML code-length on a Riemannian manifold data space Under the same conditions as in Theorem \ref{Asymptotic_PC_Easy} is given as follows.
\begin{cor}[The Rm-NML Code-Length on Riemannian Data Spaces]
\label{cor_Asymptotic_PC_Easy}
Under the same conditions as in Theorem \ref{Asymptotic_PC_Easy}, the Rm-NML code-length on a Riemannian Manifold data space can be approximated as follows.
\begin{equation}
\label{NML_codelengh_4}
\begin{aligned}
  \mathcal{L}_{\mathrm{Rm-NML}}(x^n \mid \mathcal{P})&=-\log p_{\mathrm{vol}}(x^n \mid \hat{\theta}(x^n))\\
  &+\frac{D+m}{2} \log \frac{n}{2\pi} + \log \mathrm{vol(\Theta)} \\
    &+ \log \int_{\gamma\in\Gamma} \sqrt{C_{\theta}(\gamma)C_{\gamma}(\gamma)}d\gamma + o(1).
\end{aligned}
\end{equation}
\end{cor}}

\bluetext{where $\mathrm{vol}(\Theta)$ denotes the volume of the parameter space $\Theta$.} 

\begin{remark}
\label{remark_3}
\bluetext{We remark that the formulae in Theorem \ref{Asymptotic_PC_Easy} have interpretations similar to those in Euclidean space (e.g., \cite{balasubramanian2005mdl}. The first term is derived from identifiability and corresponds to the volume of a statistically significant ellipsoid around the estimator $\hat{\theta}$ in the space of probability distributions parameterized by $\theta$, where the Fisher information matrix defines the Riemannian metric. The second and third terms arise from taking the logarithm of the volume of this distribution space. Under the conditions of Theorem \ref{Asymptotic_PC_Easy}, Lemmas~3--6 show that this distribution space can be locally represented as a direct product of a parameter space endowed with the same metric as the data space and another space corresponding to the parameter $\gamma$ equipped with a metric induced by the Fisher information. The second term corresponds to the volume of the parameter space with a metric identical to that of the data space, while the third term represents the volume of the $\gamma$-parameter space endowed with the metric derived from the Fisher information. Throughout this analysis, $\mathcal{C}(\mathcal{P})$ represents the proportion of the ellipsoidal neighborhood around $\hat{\theta}$ within the entire distribution space. }

\bluetext{Next, we discuss the reasonableness of the Rm-NML results. Corollary \ref{cor_Asymptotic_PC_Easy} provides the resulting expression for the Rm-NML code-length in this setting. One of the central applications of the general NML code-length is model selection. From the viewpoint of statistical learning theory, models with excessively large complexity are undesirable, as they tend to cause overfitting.
When the Rm-NML code-length is applied to model selection, the PC derived in Theorem \ref{Asymptotic_PC_Easy} acts as a penalty term for this complexity, namely, the model size. In the result of Corollary \ref{cor_Asymptotic_PC_Easy}, the first term is the negative log-likelihood at the maximum likelihood estimator. As the model size increases, the model generally fits the data more easily, and this term therefore decreases. For the Rm-NML to function as a model selection criterion, the subsequent terms must compensate for this decrease by acting as penalties with respect to the model size.
In Corollary \ref{cor_Asymptotic_PC_Easy}, the second term, $(D + m)$, corresponds to the degrees of freedom of the model and can be regarded as an explicit penalty for the model size. From Theorem \ref{Ap1}, it follows that the third and fourth terms in Corollary \ref{cor_Asymptotic_PC_Easy} are derived from the geometric volumes of the statistical manifold. In particular, in Corollary \ref{cor_Asymptotic_PC_Easy}, the third term is the logarithm of the volume of the parameter space $\Theta$ for $\theta$, and it functions as a penalty term that accounts for the increase in model size with respect to $\theta$. Similarly, the fourth term serves as a penalty for the increase in model size associated with the parameter $\gamma$. From these observations, we conclude that the Rm-NML code-length is consistent with the requirements imposed by statistical learning theory and can serve as a valid model selection criterion.}
\end{remark}
By applying Theorem \ref{Asymptotic_PC_Easy}, the PC for the R-GD on hyperbolic space can be obtained in a straightforward manner, as will be discussed later. Given the above Theorem \ref{Asymptotic_PC_Easy}, once the volume of the parameter space is known, the integration over the Riemannian manifold can be avoided.

\section{\bluetext{Example: Rm-NML for Gaussian Distributions on Hyperbolic Space}}
\label{Sec6}
\bluetext{As a concrete example of the proposed Rm-NML, we derive the Rm-NML code-length for the Gaussian distribution defined on hyperbolic space. Hyperbolic space has recently attracted significant attention as an embedding space for hierarchical graph data \cite{krioukov2010hyperbolic}. This is due to its exponentially expanding geometry, which is well-suited for representing tree-like hierarchical structures. In fact, previous work has demonstrated that while Euclidean embeddings may require up to 200 dimensions to represent certain hierarchical linguistic data, the same structure can be embedded in only 5 dimensions within hyperbolic space \cite{nickel2017poincare}. Therefore, hyperbolic space can be regarded as one of the most promising Riemannian manifolds for modeling complex data.}

\bluetext{In this study, we focus on the Riemannian Gaussian distribution (R-GD) as a class of probability distributions defined on Riemannian manifolds \cite{said2017gaussian}. The R-GD models probability densities based on the Riemannian distance between a data point and a mean location $\mu$ on the manifold, with a scalar parameter $\sigma$ controlling the dispersion. The R-GD is defined with respect to the volume element on the manifold, making it inherently coordinate-invariant.}

\bluetext{Specifically, we consider the Hyperbolic Gaussian Distribution (H-GD), which is an instance of the R-GD defined on hyperbolic space. Since hyperbolic space is a Riemannian symmetric space, and the H-GD belongs to the \bluetext{distance-determined-type} probability distribution family defined in this study, we can directly apply Theorem \ref{Asymptotic_PC_Easy}. Let $\mathcal{H}^D$ denote the hyperbolic space. In this study, we fix the curvature to $-1$. Where $x\in \mathcal{H}^D$, $\mu \in \mathcal{H}^D$ is the Fr\`{e}chet mean, $\sigma > 0$ is the dispersion parameter, $d_{\mathcal{H}^D}(x, \mu)$ denotes the geodesic distance between $x$ and $\mu$, and $\xi_{\mathcal{H}^D}(\sigma)$ is the normalizing constant.
The H-GD, which is the adaptation of R-GD to hyperbolic space, is then defined as follows.}

\bluetext{
\begin{equation}
\label{H_GD2}
    p_{\mathrm{vol}, \mathrm{H-GD}}(x \mid \mu, \sigma) \overset{\mathrm{def}}{=} \frac{1}{\xi(\sigma)} \exp\left( -\frac{1}{2\sigma^2} d_{\mathcal{H}^D}^2(x, \mu) \right),
\end{equation}
\begin{equation}
    \xi_{\mathrm{H-GD}}(\sigma) \stackrel{\text { def }}{=} \int_{\mathcal{H}^D} \exp \left(-\frac{1}{2 \sigma^2} d_{\mathcal{H}^D}^2(x, \mu)\right) d\mathrm{vol}(x) .
\end{equation}
$\xi_{\mathrm{H-GD}}(\sigma)$ is computed as follows. Let $p_i$ be defined as $p_i=(D-1)-2 i$.
\begin{equation}
\begin{aligned}
    &\xi(\sigma)=\frac{\pi^{\frac{D}{2}}}{\Gamma\left(\frac{D}{2}\right)} \sqrt{\frac{\pi}{2}} \frac{\sigma}{2^{D-2}} \\
    &\,\,\,\times\sum_{i=0}^{D-1}(-1)^i\binom{D-1}{i} \exp \left(\frac{\sigma^2 p_i^2}{2}\right)\left(1+\operatorname{erf}\left(\frac{p_i \sigma}{\sqrt{2}}\right)\right),
\end{aligned}
\end{equation}
we define $\operatorname{erf}(x)$ as follow.
\begin{equation}
    \operatorname{erf}(x) \stackrel{\text { def }}{=} \frac{2}{\sqrt{\pi}} \int_0^x \exp \left(-t^2\right) d t.
\end{equation}
The goal of this section is to provide a formula to compute the PC and NML of the H-GD defined above. However, where a distribution family contains a mean parameter or variance parameter as in the (Euclidean) Gaussian distribution and the H-GD, the naive NML does not exist and a modified version of NML has been used. In the following section, we explain the necessary modification, which corresponds to a restriction of data space, clarifying why such restriction is essentially necessary.
\subsection{Restriction of the data space}
The H-GD has a mean parameter and a variance parameter, as the (Euclidean) Gaussian distribution do. For the Gaussian distribution, it has been known that the NML in the naive sense, characterized as the solution to the minimax regret problem (18), does not exist. The non-existence of the naive NML implies that the original minimax regret problem was overambitious in the sense that the ``max'' part of the minimax problem considers all the possible data sequences. To relax the naive and infeasible minimax regret problem into a feasible problem, the restriction of the data space in the ``max'' part of the minmax regret problem has been widely adopted in existing work handling the NML of the (Euclidean) Gaussian distribution, e.g., \cite{rissanen2002mdl, hirai2013efficient, hirai2012normalized, suzuki2021fourier}. Note that the smaller set the ``max'' part considers, the easier and more feasible a minimax problem. Following the previous work, we also apply the same type of the restriction to formulate the H-GD's NML. Specifically, we consider the following data space restriction. When restricting the data space, we place the origin $o_{\mathcal{H^D}}$ in the hyperbolic space.
\begin{equation}
\label{restriction_1}
\begin{aligned}
& \mathcal{X}^n(R, \sigma_{\mathrm{min}}, \sigma_{\mathrm{max}}) \\
& \stackrel{\text { def }}{=}\left\{x^n:d_{\mathcal{H}^D}(\hat{\mu}(x^n), o_{\mathcal{H^D}}) \leq R,\, \sigma_{\mathrm{min}} \leq \hat{\sigma}(x^n) \leq \sigma_{\mathrm{max}}\right\}.
\end{aligned}
\end{equation}
Note that while the above restriction is formally applied to the data space, we can also conceptually view it as a restriction on the parameter space, as the restriction is defined as the location of the maximum likelihood estimator in the parameter space. 
Under the restriction imposed by Equation (\ref{restriction_1}), the minimax regret problem is modified to the following:
\begin{equation}
\label{min_max_RmNML2}
\begin{aligned}
    &p_{\mathrm{Rm\text{-}NML}}(\cdot)\\
    &=\underset{q_{\mathrm{vol}}}{\operatorname{argmin}} \max _{x^n\in \mathcal{X}^n(R, \sigma_{\mathrm{min}}, \sigma_{\mathrm{max}})}\\
    &\qquad\left\{-\log q_{\mathrm{vol}}(x^n)-\min _\theta(-\log p_{\mathrm{vol}}(x^n \mid \theta))\right\}.
\end{aligned}
\end{equation}
As the solution to the above modified minimax regret, we obtain the following modified version of Rm-NML.
\begin{equation}
\label{LNML_rmnml}
\begin{aligned}
&p_{\mathrm{Rm-NML}}(x^n) \\
&=
\frac{
p\!\left(x^n \mid \hat{\theta}(x^n)\right)\, \mathbf{1}_{\mathcal{X}^n(R, \sigma_{\mathrm{min}}, \sigma_{\mathrm{max}})}(\hat{\theta}(x^n))
}{
\int
p\!\left(y^n \mid \hat{\theta}(y^n)\right)\, \mathbf{1}_{\mathcal{X}^n(R, \sigma_{\mathrm{min}}, \sigma_{\mathrm{max}})}(\hat{\theta}(x^n))\, d(y^n)
}.
\end{aligned}
\end{equation}
In this section, as far as we discuss the H-GD, the Rm-NML refers to the above modified version. Also, the PC refers to the denominator of the modified Rm-NML.
Here, we can see that the data space restriction can also be viewed as a restriction in the integral domain in the PC. 
We can also see that the restriction is consistent to our Theorem \ref{Asymptotic_PC_Easy} in the sense that it leads to satisfy the compactness condition (Condition
1) and the finiteness of the integral (Condition 3).
Nevertheless, we emphasize that the necessity of the data space restriction comes from the NML itself, not indicating a flaw of our computation theorems.}

\bluetext{
\begin{remark}[About the choice of R]
    A practical method for selecting the parameter space restriction $R$ is to determine it empirically based on the observed data. In particular, a simple and effective heuristic is to set $R$ equal to the maximum distance from the origin among the observed data points, as suggested in Corollary \ref{COR_NML_codelengh_5}. This approach provides a straightforward way to ensure that the NML code-length is well defined without requiring additional theoretical adjustments.
    Note that although methods based on renormalized normalized maximum likelihood (RNML) \cite{rissanen2002mdl, hirai2012normalized} have been proposed to alleviate the dependency of the code-length on the choice of R in the literature of model selection regarding the (Euclidean) Gaussian distribution, extending the RNML to the Riemannian case is nontrivial and remains future work.
\end{remark}}

\begin{comment}

\subsection{Problem definition}
We consider a given data sequence $x^n$ located in the D-dimensional hyperbolic space $\mathcal{H}^D$. Hyperbolic space $\mathcal{H}^D$ is a Riemannian manifold of constant negative curvature, equipped with the following inner product structure. In this study, we fix the curvature to $-1$. We also consider the Lorentz model as a hyperbolic model. The data space is restricted to $\mathcal{X}^n(R, \sigma_{\mathrm{min}}, \sigma_{\mathrm{max}})$. We compute the value of the Rm-NML in Equation (\ref{LNML_rmnml}) under this setting, in particular focusing on the value of its PC.
\end{comment}

\bluetext{
\subsection{NML computation formula for H-GD}
In this subsection, we provide the computation formula for the Rm-NML and PC of the H-GD. We remark that the Rm-NML and PC here are the data space restricted version introduced in the previous subsection. Specifically, the data space is restricted to $\mathcal{X}^n(R, \sigma_{\mathrm{min}}, \sigma_{\mathrm{max}})$. The following is our asymptotic PC computation formula for H-GD obtained as a corollary of Theorem \ref{Asymptotic_PC_Easy}.}

\begin{cor}[Computation of the PC for Riemannian Gaussian Distributions on Hyperbolic Space via Asymptotic Analysis]
\label{cor1}
When $n$ data points are distributed on the D-dimensional hyperbolic space 
$\mathcal{H}^D$ , the PC of the R-GD can be asymptotically approximated as follows. The data space is restricted to \( x^n \in \mathcal{X}^n(R, \sigma_{\mathrm{min}}, \sigma_{\mathrm{max}}) \). For notational convenience, we define $B(\sigma)$ as follows. 
\begin{equation}
    B(\sigma) \stackrel{\text { def }}{=} \sqrt{\frac{1}{\sigma}\left\{\sigma \frac{\xi^{\prime \prime}(\sigma)}{\xi(\sigma)}-\sigma\left(\frac{\xi^{\prime}(\sigma)}{\xi(\sigma)}\right)^2+3 \frac{\xi^{\prime}(\sigma)}{\xi(\sigma)}\right\}}.
\end{equation}
Let $\mathrm{V}_{\mathcal{H}^D}(R)$ denote the volume of a ball of radius $R$ in D-dimensional hyperbolic space. The PC ($\mathcal{C}(\mathcal{P})$) in Equation (\ref{PC_label2}) for the H-GD on hyperbolic space satisfies the following. 
\begin{equation}
\label{RmNML_HGD_PC}
    \begin{aligned}
&\log \mathcal{C}(\mathcal{P})= \frac{D+1}{2} \log \frac{n}{2 \pi}+\log \mathrm{V}_{\mathcal{H}^D}(R) \\
& +\log \int_{\sigma_{\mathrm{min}}}^{\sigma_{\mathrm{max}}}\left(\frac{1}{D \sigma^2} \frac{\xi^{\prime}(\sigma)}{\xi(\sigma)}\right)^{\frac{D}{2}} B(\sigma) d \sigma+o(1).
\end{aligned}
\end{equation}
See Appendix \ref{Ap4} for the proof.
\end{cor}

\bluetext{By Corollary \ref{cor1}, the Rm-NML code-length for the R-GD on hyperbolic space is computed as follows.
\begin{cor}[Computation of the Rm-NML code-length for Riemannian Gaussian Distributions on Hyperbolic Space]
Under the same conditions as in Corollary \ref{cor1}, the Rm-NML code-length for the R-GD on hyperbolic space is computed as follows.
\label{COR_NML_codelengh_5}
\begin{equation}
\label{NML_codelengh_7}
\begin{aligned}
  \mathcal{L}_{\mathrm{Rm-NML}}&(x^n \mid \mathcal{P})=\log \xi(\hat{\sigma}) +\sum_{i=1}^{n}\frac{1}{2\hat{\sigma}^2} d^2(x_i, \hat{\mu})\\
  &+\frac{D+1}{2} \log \frac{n}{2 \pi}+\log \mathrm{V}_{\mathcal{H}^D}(R) \\
& +\log\int_{\sigma_{\mathrm{min}}}^{\sigma_{\mathrm{max}}}\left(\frac{1}{D \sigma^2} \frac{\xi^{\prime}(\sigma)}{\xi(\sigma)}\right)^{\frac{D}{2}} B(\sigma) d \sigma+o(1).
\end{aligned}
\end{equation}
\end{cor}}
Since the integral with respect to $\sigma$ in Corollary \ref{cor1} and Corollary \ref{COR_NML_codelengh_5} is analytically intractable, its value is computed using numerical integration.

\bluetext{Each term appearing in Corollary~\ref{cor1} and Corollary~\ref{COR_NML_codelengh_5} admits an interpretation analogous to that given in Remark~\ref{remark_3}. From the viewpoint of statistical learning theory, models with excessively large complexity are undesirable, as they tend to cause overfitting. One of the central applications of NML is model selection, where the PC is known to function as a penalty term that accounts for the size of the model.
With respect to the radius \(R\), the model size can be regarded as geometrically monotone increasing. When the dimension \(D\) differs, the underlying data domains themselves are different, and hence a direct comparison is not straightforward. Nevertheless, as suggested by the theory underlying criteria such as MDL \cite{rissanen2002mdl}, Akaike Information Criterion (AIC) \cite{akaike2003new} and Bayesian Information Criterion (BIC) \cite{schwarz1978estimating}, it is generally expected that the model size increases monotonically with respect to the parameter dimension. Equation~(\ref{RmNML_HGD_PC}) is monotone increasing in both \(D\) and \(R\), and is therefore consistent with these requirements arising from statistical learning theory.}

The determinant of the Fisher information matrix with respect to the mean parameter $\mu$ becomes a constant due to the coordinate transformation into an orthonormal basis and isometric transformation at each point. Let this constant be denoted as $I_{\zeta_{\nu}}(\nu)$. The resulting expression for the R-GD on hyperbolic space is given below. Here, $\xi^{\prime}(\sigma)$ denotes the derivative of the normalization factor $\xi(\sigma)$ with respect to $\sigma$, and $E$ is the D-dimensional identity matrix.
\begin{equation}
    I_{\zeta_{\nu}}(\nu)=\frac{1}{D \sigma} \frac{\xi^{\prime}(\sigma)}{\xi(\sigma)} E,
\end{equation}
\begin{equation}
    \log \sqrt{\left|I_{\zeta_{\nu}}(\nu)\right|}=\frac{D}{2} \log \left\{\frac{1}{D \sigma} \frac{\xi^{\prime}(\sigma)}{\xi(\sigma)}\right\} .
\end{equation}
The Fisher information $I_{\sigma}$ with respect to the scale parameter $\sigma$ for the R-GD on hyperbolic space is given as follows.
\begin{equation}
    I_\sigma=\frac{1}{\sigma}\left\{\sigma \frac{\xi^{\prime \prime}(\sigma)}{\xi(\sigma)}-\sigma\left(\frac{\xi^{\prime}(\sigma)}{\xi(\sigma)}\right)^2+3 \frac{\xi^{\prime}(\sigma)}{\xi(\sigma)}\right\}.
\end{equation}
$\mathrm{V}_{\mathcal{H}^D}(R)$ denote the volume of a ball of radius $R$ in D-dimensional hyperbolic space. This volume is given by the following expression \cite{ratcliffe2006foundations}.
\begin{equation}
\label{vol_spere_H}
\begin{aligned}
    &\mathrm{V}_{\mathcal{H}^D}(R) \\
    &=\left\{\frac{\pi^{\frac{D}{2}}}{2^{D-2} \Gamma\left(\frac{D}{2}\right)} \sum_{i=0}^{D-1}(-1)^i \frac{\exp \{(D-1-2 i) R\}-1}{D-1-2 i}\right\} .
\end{aligned}
\end{equation}

\bluetext{In Theorem \ref{Asymptotic_PC_Easy}, it is shown that for distance-determined-type families on Riemannian symmetric spaces, the PC can be computed once the following quantities are known: the volume of the parameter space of parameters that lie on the same Riemannian manifold as the data points, the determinant of the Fisher information matrix with respect to those parameters, and the determinant of the Fisher information matrix with respect to the remaining parameters. In the example of the hyperbolic Gaussian distribution, the integration range of the parameter space is restricted to the region where the distance from the origin is less than or equal to $R$. Therefore, if the volume of a ball in hyperbolic space is known, the volume of the parameter space can be computed. Since a closed-form formula for this volume exists, the calculation is straightforward. Moreover, in this example, Theorem \ref{Asymptotic_PC_Easy} implies that the determinant of the Fisher information matrix can be computed in a factorized manner. As a result, it is not necessary to evaluate mixed derivative terms involving both $\theta$ and $\sigma$, which significantly simplifies the derivation of the determinant. Corollary \ref{COR_NML_codelengh_5} is a concrete example of Theorem \ref{Asymptotic_PC_Easy}, and it supports the claim that the computation is indeed simplified by Theorem \ref{Asymptotic_PC_Easy}. Moreover, by computing the parametric complexity, we can quantify how much additional code-length is incurred when encoding R-GD on hyperbolic space using Rm-NML code-length.}

\section{Conclusion}
\label{Sec7}
In this study, we have presented, for the first time, a theoretical framework for calculating the NML code-length on Riemannian data spaces in a coordinate-invariant manner. The proposed Rm-NML formulation allows for the computation of code-lengths that are invariant under coordinate transformations, thereby ensuring geometric consistency in statistical modeling on curved spaces.

By employing Rm-NML, it becomes possible to conduct regret analysis based on the MDL principle on Riemannian manifolds, just as has been done using NML in Euclidean spaces, while preserving geometric consistency. The computation of the PC, however, is generally challenging, and it was previously unclear whether asymptotic approximation techniques—originally developed in Euclidean settings—could be extended to Riemannian data spaces. In this study, we derive asymptotic approximations of PC on Riemannian manifolds and demonstrate that both the approximated expressions and their regularity conditions are preserved under smooth coordinate transformations. This provides a coordinate-invariant method for computing the asymptotic form of PC on Riemannian data spaces. Nevertheless, even with such approximations, computing PC remains difficult due to the extension of the data space to general Riemannian manifolds.

To address this, we focus on data spaces that exhibit symmetric structures, which have received increasing attention in recent studies. Specifically, we develop a simplified approach to computing the PC for probability distributions defined based on the geodesic structure of Riemannian symmetric spaces. This method allows us to avoid integration over the data space when the volume of the parameter space is known, thereby significantly reducing the computational complexity of PC. Furthermore, this framework is applicable to Riemannian manifolds that have attracted substantial interest in recent years. As a concrete example, we provide the asymptotic form of the PC for R-GD defined on hyperbolic space.

By utilizing the Rm-NML, it becomes possible to apply the MDL principle on data spaces endowed with Riemannian geometry, while preserving geometric consistency. As a result, various applications of regret-based analysis using NML can now be extended to Riemannian manifolds.

\appendices
\section{Information Theoretic Justification of Code-Lengths for Continuous Probability Distributions}
\label{Information_Theoretic_Justification}
\bluetext{In this section, we give the proofs of Theorem \ref{IFJ1} and Theorem \ref{IFJ2}, which establish an information-theoretic justification for code-lengths in continuous probability distributions. We first prove the existence result stated in Theorem \ref{IFJ1}.
\begin{proof}
For a prefix code with code-lengths $\ell_S$ to exist, the Kraft inequality $\sum_{S \in \rho} 2^{-\ell_S} \le 1$ must be satisfied. Let us verify this condition. By definition, we have $\ell_S \ge \sup_{x \in S} \bigl( - \log_2 p(x) \bigr) - \log_2 \operatorname{vol}(S)$.
\begin{equation}
2^{-\ell_S} \leq 2^{-\left(\sup _{x \in S}\left(-\log _2 p(x)\right)-\log _2 \operatorname{vol}(S)\right)}.
\end{equation}
Transform the right-hand side.
\begin{equation}
\begin{aligned}
2^{\log _2 \operatorname{vol}(S)} \cdot 2^{-\sup _{x \in S}\left(-\log _2 p(x)\right)} & =\operatorname{vol}(S) \cdot 2^{\inf _{x \in S}\left(\log _2 p(x)\right)} \\
& =\operatorname{vol}(S) \cdot \inf _{x \in S}(p(x)).
\end{aligned}
\end{equation}
Therefore, we can conclude the following.
\begin{equation}
\sum_{S \in \rho} 2^{-\ell_S} \leq \sum_{S \in \rho} \operatorname{vol}(S) \inf _{x \in S}(p(x)).
\end{equation}
Here, since $\inf _{x \in S}(p(x)) \cdot \operatorname{vol}(S) \leq \int_S p(x) d \mu(x)$, the following holds.
\begin{equation}
\sum_{S \in \rho} \operatorname{vol}(S) \inf _{x \in S}(p(x)) \leq \sum_{S \in \rho} \int_S p(x) d \mu(x)=\int_{\mathcal{X}} p(x) d \mu(x).
\end{equation}
Since $p(x)$ is a probability density function, $\int_{\mathcal{X}} p(x)\, d\mu(x) = 1$, and hence
$\sum_{s \in \rho} 2^{-\ell_s} \le 1$ holds. This implies the existence of a prefix code with codeword
lengths $\ell_s$.
\end{proof}}
\bluetext{Next, we show that the coding scheme described in the previous section is optimal
in terms of the average code-length.
Let the probability that $x$ belongs to a region $S$ be given by
\[
P(S) = \int_S p(x)\, d\mu(x).
\]
Therefore, the average code-length required to specify the region $S$ to which $x$
belongs is written as
\[
\bar{L} = \sum_{S \in \rho} P(S)\, \ell^{*}_S,
\]
where $\ell^{*}_S$ denotes the codeword length assigned to region $S$.
\begin{proof}
By the source coding theorem, the average code-length of any instantaneous
decodable code (including prefix codes) cannot be smaller than the entropy.
That is,
\[
\bar{L} \ge H(\{P(S)\}_{S \in \rho})
= - \sum_{S \in \rho} P(S) \log_2 P(S).
\]
We now show that the average code-length is bounded from below by $L_{\mathrm{lower}}$.
\begin{align*}
L_{\mathrm{lower}}
&= \mathbb{E}_S\!\left[ \inf_{x \in S} ( -\log_2 p(x) ) - \log_2 \operatorname{vol}(S) \right] \\
&=\sum_{S \in \rho} P(S)
\left( \inf_{x \in S} \bigl( -\log_2 p(x) \bigr) - \log_2 \operatorname{vol}(S) \right).
\end{align*}
To this end, we consider the difference between $H(\{P(S)\})$ and $L_{\mathrm{lower}}$:
\begin{align*}
&H(\{P(S)\}) - L_{\mathrm{lower}}\\
&= - \sum_{S \in \rho} P(S) \log_2 P(S)\\
&\quad- \sum_{S \in \rho} P(S)
 \left( \inf_{x \in S} \bigl( -\log_2 p(x) \bigr) - \log_2 \operatorname{vol}(S) \right) \\
&= \sum_{S \in \rho} P(S)
\left[
- \log_2 P(S)
- \left( - \sup_{x \in S} \log_2 p(x) - \log_2 \operatorname{vol}(S) \right)
\right] \\
&= \sum_{S \in \rho} P(S)
\log_2 \frac{\sup_{x \in S} p(x) \cdot \operatorname{vol}(S)}{P(S)}.
\end{align*}
Since
\[
P(S) = \int_S p(x)\, d\mu(x)
\le \sup_{x \in S} p(x) \cdot \operatorname{vol}(S),
\]
it follows that
\[
\frac{\sup_{x \in S} p(x) \cdot \operatorname{vol}(S)}{P(S)} \ge 1,
\]
and hence each term in the sum is nonnegative.
Therefore,
\[
H(\{P(S)\}) - L_{\mathrm{lower}} \ge 0.
\]
Consequently,
\[
\bar{L} \ge H(\{P(S)\}) \ge L_{\mathrm{lower}},
\]
which shows that the average code-length cannot be smaller than
$\mathbb{E}_S\!\left[ \inf_{x \in S} ( -\log_2 p(x) ) - \log_2 \operatorname{vol}(S) \right]$.
\end{proof}
By establishing these two theorems, the Rm-NML defined in Definition \ref{Rm_NML_def22} can be interpreted as a valid code-length.}

\section{Asymptotic approximation formula for PC on Riemannian manifolds}
\label{Ap1}
We provide a proof of the asymptotic approximation of the PC on a Riemannian manifold, which is derived in this study and presented as Theorem \ref{Asymptotic_Rm}. 
\begin{proof}
    In Theorem \ref{PC_Asy}, the parameter is denoted by $\theta$, while in Theorem \ref{Asymptotic_Rm}, it is denoted by $\phi(\theta)$. This distinction in notation is used to emphasize the claim of coordinate invariance. Even in Theorem \ref{PC_Asy}, integration and differentiation are carried out with respect to the specified coordinate system. Theorem \ref{PC_Asy} is proven, following Rissanen (1996), using a hyper-rectangular partition of the parameter space. Although such a construction cannot be explicitly employed when dealing with Riemannian manifolds, the second term on the right-hand side of the asymptotic approximation in Theorem \ref{Asymptotic_Rm} involves integration with respect to the coordinate representation of the parameter, rather than with respect to the volume element. Therefore, the proof technique based on hyper-rectangles remains valid. The key issue lies in the definition of the Fisher information matrix.

    The definitions of the Fisher information matrix in Theorem \ref{PC_Asy} and Theorem \ref{Asymptotic_Rm} yield the same numerical values. As previously discussed, there is no issue regarding differentiation. The key distinction lies in the definition of the probability density function: in Theorem \ref{PC_Asy}, it is defined with respect to infinitesimal coordinate volume elements, whereas in Theorem \ref{Asymptotic_Rm}, it is defined with respect to the Riemannian volume element. Nevertheless, as shown in the following derivation, the resulting Fisher information matrices are equivalent.

    \begin{equation}
    \begin{aligned}
        \frac{\partial^2 \log p(x^n\mid \theta)}{\partial \theta \partial \theta^\top}&=\frac{\partial^2 \log p_{\mathrm{vol}}(x^n\mid \theta)\sqrt{\det g_{\mathcal{M}}(x)}}{\partial \theta \partial \theta^\top} \\
        &=\frac{\partial^2 \log p_{\mathrm{vol}}(x^n\mid \theta)}{\partial \theta \partial \theta^\top}+\frac{\partial^2 \log \sqrt{\det g_{\mathcal{M}}(x)}}{\partial \theta \partial \theta^\top}\\
        &=\frac{\partial^2 \log p_{\mathrm{vol}}(x^n\mid \theta)}{\partial \theta \partial \theta^\top}.
    \end{aligned}
    \end{equation}

    As demonstrated above, the Fisher information matrices yield identical values. Therefore, Theorem \ref{Asymptotic_Rm} holds in the same manner as Theorem \ref{PC_Asy}.
\end{proof}

\section{Coordinate Invariance of the Conditions for the Asymptotic Approximation}
\label{Ap2}
We prove that the five regularity conditions stated in Theorem \ref{Asymptotic_Rm} are preserved under any twice continuously differentiable coordinate transformation. Conditions 2, which concerns the continuity of the Fisher information matrix, and 3, which requires the finiteness of the resulting PC value, are considered to be self-evident under coordinate transformations. Here, we assume that all conditions are satisfied with respect to a coordinate system $\phi$, and consider a transformation to another coordinate system $\psi$ via a twice continuously differentiable change of coordinates.
\subsection{Coordinate Invariance of Condition 1}
Assume that $\Theta$ is compact, and that there exist positive constants $c_1, c_2$ such that for all $\theta\in\Theta$, $0 < c_1 \leq |I_\phi(\theta)| \leq c_2 < \infty$. Let $\psi$ be the coordinate system obtained from $\phi$ via a twice continuously differentiable transformation. Then, we aim to prove that $\Theta$ remains compact, and that there exist positive constants $c_1^{\prime}, c_2^{\prime}$ such that for all $\theta\in\Theta$, $0 < c_1^{\prime} \leq |I_\psi(\theta)| \leq c_2^{\prime} < \infty$.
\begin{proof}
\begin{equation}
\left|I_\phi(\theta)\right|=\left(\det \frac{\partial \psi}{\partial \phi}(\theta)\right)^2\left|I_{\psi}(\theta)\right| .
\end{equation}
Since $\det \frac{\partial \psi}{\partial \phi}(\theta)$ is a continuous function with respect to $\theta$, and $\Theta$ is compact, there exists a maximum value $b_{max}$. From Condition 4, since $I_{\psi}(\theta)<C_0^{\prime}$, $I_{\psi}$ is bounded and does not diverge. Moreover, from Condition 1, $\det \frac{\partial \psi}{\partial \phi}(\theta)$ is non-zero.
\begin{equation}
b_{\min } \leq\det \frac{\partial \psi}{\partial \phi}(\theta)\leq b_{\max},
\end{equation}
\begin{equation}
\begin{aligned}
0<c_1 &\leq\left|I_\phi(\theta)\right| \leq c_2<\infty \\
&\Rightarrow 0<c_1 \leq\left(\det \frac{\partial \psi}{\partial \phi}(\theta)\right)^2\left|I_{\psi}(\theta)\right| \leq c_2<\infty,
\end{aligned}
\end{equation}
\begin{equation}
\begin{aligned}
0<c_1 &\leq\left(\det \frac{\partial \psi}{\partial \phi}(\theta)\right)^2\left|I_{\psi}(\theta)\right| \leq c_2<\infty\\ &\Rightarrow 0<\frac{c_1}{b_{\max }^2} \leq\left|I_{\psi}(\theta)\right| \leq \frac{c_2}{b_{\min }^2}<\infty.
\end{aligned}
\end{equation}
From the above, it follows that $c_2^{\prime}=\frac{c1}{b_{max}^2}$, $c_2^{\prime}=\frac{c2}{b_{min}^2}$, and hence $c_1^{\prime}, c_2^{\prime}$ exist. 
\end{proof}

\subsection{Coordinate Invariance of Condition 5}
To verify whether the following condition remains valid even under coordinate transformations that are twice continuously differentiable, we consider coordinate systems obtained via such transformations. Suppose there exists a finite positive definite matrix \( C_0 \) such that the following inequality holds:
\begin{align}
    I_{\phi}(x^n, \theta) < C_0.
\end{align}

Let \(\psi\) be a coordinate system obtained by applying a twice continuously differentiable coordinate transformation to the coordinate system \(\phi\). Suppose there exists a finite positive definite matrix \( C_0' \) such that the following also holds:
\begin{align}
    I_{\psi}(x^n, \theta) < C_0'.
\end{align}
\begin{proof}
The Jacobian matrix
\[
    \frac{\partial \phi}{\partial \psi}(\theta)
\]
is a matrix-valued function that is continuous with respect to \(\theta\), and the parameter space \(\Theta\) is compact. Therefore, the matrix norm attains its maximum value:
\begin{align}
\label{123}
    \left[ \frac{\partial \phi}{\partial \psi}(\theta) \right] < C_{\phi, \psi}.
\end{align}

Then, the Fisher information matrix under the transformed coordinate system \(\psi\) can be written as:
\begin{align}
    I_{\psi}(x^n, \theta) = \left[ \frac{\partial \phi}{\partial \psi}(\theta) \right]^\top 
    I_{\phi}(x^n, \theta)
    \left[ \frac{\partial \phi}{\partial \psi}(\theta) \right]. 
\end{align}

Using the inequality in Equation (\ref{123}) and the assumption on \(I_{\phi}(x^n, \theta)\), we obtain:
\begin{align}
    I_{\psi}(x^n, \theta) 
    &< C_{\phi, \psi}^\top I_{\phi}(x^n, \theta) C_{\phi, \psi} \notag \\
    &< C_{\phi, \psi}^\top C_0 C_{\phi, \psi}.
\end{align}

Therefore, letting
\begin{align}
    C_0' = C_{\phi, \psi}^\top C_0 C_{\phi, \psi},
\end{align}
we conclude that such a \( C_0' \) exists. 
\end{proof}
\subsection{Coordinate Invariance of Condition 4}
We verify that the uniform asymptotic normality as defined in Definition~\ref{Condition4} is preserved under twice continuously differentiable coordinate transformations. Since the proof of the coordinate invariance of Condition~4 is quite lengthy, we present only an outline of the proof.

We begin by establishing the following two lemmas.
\begin{lem}
Suppose the function \( \nabla_{\phi} \log p(x^n \mid \theta) \) is twice continuously differentiable with respect to \( \phi(\theta) \), and that the maximum likelihood estimator \( \hat{\theta} \) satisfies
\[
\nabla_{\phi} \log p\left(x^n \mid \phi(\hat{\theta}(x^n))\right) = 0.
\]
If the following asymptotic normality holds for some \( \theta_0 \in \Theta \),
\begin{align}
\sqrt{n} \left( \phi\left( \hat{\theta}(x^n) \right) - \phi(\theta) \right)
\xrightarrow{d} \mathcal{N}\left( 0,\, I_{\phi}(\theta)^{-1} \right), 
\end{align}
then we have
\begin{align}
\frac{1}{\sqrt{n}} \nabla_{\phi} \log P\left(x^n \mid \phi(\theta_0) \right)
\xrightarrow{d} \mathcal{N}\left( 0,\, I_{\phi}(\theta_0) \right). 
\end{align}
\end{lem}
\begin{lem}
Assume that $\nabla_\phi \log p(x^n \mid \phi(\theta))$ is twice continuously differentiable with respect to $\phi(\theta)$, and that the maximum likelihood estimator $\hat{\theta}=\hat{\theta}(x^n)$ satisfies 
\[
0 = \nabla_\phi \log p\left(x^n \mid \phi\left(\hat{\theta}(x^n)\right)\right).
\]
When the regularity conditions for asymptotic normality hold, the following result also holds for the reparameterized coordinate system $\psi(\theta)$, which is obtained by a smooth and invertible transformation $\psi$ of the parameter $\theta$:
\[
\sqrt{n} \left( \psi\left( \hat{\theta}(x^n) \right) - \psi(\theta) \right) \xrightarrow{d} \mathcal{N}\left(0, I_\psi(\theta)^{-1} \right). 
\]
\end{lem}
Both of the above results can be derived using the first-order Taylor expansion, the law of large numbers, and Slutsky's theorem. Using these lemmas, we establish the coordinate invariance of Condition 4.
The term bounded by $\epsilon$ in Definition \ref{Condition4} is decomposed as follows, and by considering an open ball whose radius increases with $n$, the desired result is obtained.
\begin{equation}
B \stackrel{\text { def }}{=} \phi^{-1}\left(\left(\frac{1}{\sqrt{n}} A_\phi+\phi(\theta)\right) \cap \phi(\Theta)\right) .
\end{equation}
\begin{equation}
\begin{aligned}
    &\left| p\left( \sqrt{n} \left( \psi\left( \hat{\theta}(x^n) \right) - \psi(\theta) \right) \in A_\psi \right) - p\left( X \in A_\psi \right) \right| \\
&< | p\left( \hat{\theta}(x^n) \in B \right) - p\left( X_\phi \in \sqrt{n}(\phi(B) - \phi(\theta)) \right) \\
&- p\left( X_\phi \in A_\phi \setminus \sqrt{n}(\phi(\Theta) - \phi(\theta)) \right) | \\
&+| p\left( X_\phi \in \sqrt{n}(\phi(B) - \phi(\theta)) \right) \\
&- p\left( X_\psi \in \sqrt{n}(\psi(B) - \psi(\theta)) \right)| \\
&+ \left|p\left( X_\phi \in A_\phi \setminus \sqrt{n}(\phi(\Theta) - \phi(\theta)) \right)\right| \\
&+ \left| p\left( X_\psi \in A_\psi \setminus \sqrt{n}(\psi(\Theta) - \psi(\theta)) \right) \right|.
\end{aligned}
\end{equation}

\section{A simple method for computing asymptotic approximations of PC}
\label{Ap3}
To prove Theorem \ref{Asymptotic_PC_Easy}, we establish the following four lemmas. 
\begin{lem}
    Under the same setting as Theorem \ref{Asymptotic_PC_Easy}, let $\mathcal{M}$ be a Riemannian manifold and $\theta\in\mathcal{M}$ be an arbitrary point. The coordinate transformation of the determinant of the Fisher information matrix $I_{\phi}(\theta, \gamma)$, defined with respect to a global coordinate system $\phi$ and a locally orthonormal coordinate system $\zeta^\theta$ at $\theta$, satisfies the following expression.
    \begin{equation}
        |I_\phi(\theta, \gamma)| = |I_{\zeta^\theta}(\theta, \gamma)|  \det g_{\mathcal{M}}(x).
    \end{equation}
\end{lem}
\begin{lem}
    Under the same setting as Theorem \ref{Asymptotic_PC_Easy}, consider a coordinate system $\zeta^\nu$ defined by an orthonormal basis at an arbitrary reference point $\nu \in \mathcal{M}$ on the Riemannian manifold $\mathcal{M}$.  Then, the determinant of the Fisher information matrix $I_{\zeta^\nu}(\nu, \gamma)$, evaluated at $\nu$ for fixed $\gamma$, is independent of the choice of $\nu$.
    \begin{equation}
        \exists C \in \mathbb{R}, \ \forall \nu \in \mathcal{M}, \ |I_\psi(\nu, \gamma)| = C_\theta(\gamma).
    \end{equation}
\end{lem}
\begin{lem}
    Under the same setting as Theorem \ref{Asymptotic_PC_Easy}, the Fisher information matrix $I(\phi(\theta), \gamma)$ becomes block-diagonal and can be expressed as follows:
    \begin{equation}
        I(\phi(\theta), \gamma) = 
\begin{bmatrix}
I_\phi(\theta, \gamma) & 0 \\
0 & I_\gamma(\theta, \gamma)
\end{bmatrix}.
    \end{equation}
\end{lem}
\begin{lem}
    Under the same setting as Theorem \ref{Asymptotic_PC_Easy}, the Fisher information matrix $I_{\gamma}(\nu, \gamma)$, with respect to $\gamma$, is independent of $\theta$, and satisfies:
    \begin{equation}
        \forall v_1, v_2 \in \mathcal{M}, \quad I_\gamma(v_1, \gamma) = I_\gamma(v_2, \gamma).
    \end{equation}
\end{lem}
Lemmas 3, 5, and 6 can be proven by relatively straightforward variable transformations. In contrast, the proof of Lemma 4 relies on the use of isometric transformations. Theorem \ref{Asymptotic_PC_Easy} considers a Riemannian symmetric space, which is also a Riemannian homogeneous space. In a Riemannian homogeneous space, isometries exist transitively; that is, for any two points, there exists an isometry that maps one to the other. An isometry is a mapping that preserves distances. Lemma 4 is proven by considering orthonormal bases at two points that are related via such an isometry, and by computing the determinant of the Fisher information matrix at each point.

\section{Computation of the PC for H-GD via Asymptotic Analysis}
\label{Ap4}
We derive the PC value of R-GD on hyperbolic space.
By applying Theorem \ref{Asymptotic_PC_Easy}, the integral over hyperbolic space can be avoided.
Thus, the quantities that actually need to be evaluated are the volume of the parameter space $\Theta$ and the constants $C_{\theta}(\gamma)$ and $C_{\gamma}(\gamma)$ appearing in Theorem \ref{Asymptotic_PC_Easy}. The volume of the parameter space can be computed as the volume of a geodesic ball in hyperbolic space, restricted to a radius $R$, and is given by Equation (\ref{vol_spere_H}).
There exist multiple equivalent models for representing hyperbolic space. In this work, we adopt the Lorentz model with the following metric and proceed with the calculations accordingly.
\begin{equation}
d_s^2=-d \phi_0^2+d \phi_1^2+\cdots d \phi_D^2.
\end{equation}
The polar coordinate representation in the Lorentz model is given as follows.
\begin{equation}
(\phi_1, \cdots, \phi_D)^\top=\left[\begin{array}{c}
\sinh r \cos \theta_1 \\
\vdots \\
\sinh r \sin \theta_1 \cdots \sin \theta_{D-2} \cos \theta_{D-1} \\
\sinh r \sin \theta_1 \cdots \sin \theta_{D-2} \sin \theta_{D-1}
\end{array}\right].
\end{equation}
Theorem \ref{Asymptotic_PC_Easy} asserts that the value of $C_{\theta}(\gamma)$ is constant and does not depend on the choice of representative point $\nu$. As the representative point, we consider the origin $o_h$ in the Lorentz model, where the coordinate index satisfies $(\phi_0, \phi_1, \cdots, \phi_D)=(1, 0, \cdots, 0)$. 
As a result of the computation, the following expression is obtained.
\begin{equation}
{\left[\frac{\partial d^2\left(x, \mu\right)}{\partial \mu_i}\right]_{\mu=o_h} }  =-2 \frac{r}{\sinh r} x , 
\end{equation}
\begin{equation}
\label{int_1}
\begin{aligned}
&I_\phi\left(o_h, \sigma\right)\\
&=\frac{1}{\sigma^4} \int_{x \in \mathcal{H}^D} \frac{1}{\xi(\sigma)} \exp \left(-\frac{r^2}{2 \sigma^2} \right)\left(\frac{r}{\sinh r}\right)^2 x x^{\top} d \operatorname{vol}(x).
\end{aligned}
\end{equation}
Due to symmetry, it is sufficient to compute a single diagonal component of $I_\phi\left(o_h, \sigma\right)$. In the following, we consider the (1,1)-component and denote the diagonal entry by $I_\phi\left(o_h, \sigma\right)_{i,i}$. By rewriting Equation (\ref{int_1}) in terms of the polar coordinate representation, the constant $I_\phi\left(o_h, \sigma\right)$ can be decomposed as follows.
\begin{equation}
\begin{aligned}
& I_\phi\left(o_h, \sigma\right)_{i, i} \\
& =\frac{1}{\sigma^4} \int \frac{1}{\xi(\sigma)} \exp \left(-\frac{1}{2 \sigma^2} r^2\right) r^2 \sinh ^{D-1} r d r \\
&\quad \times \int \cos \theta_1^2 \sin ^{D-2} \theta_1 \cdots \sin \theta_{D-2} d \theta_1 \cdots d \theta_{D-1} .
\end{aligned}
\end{equation}
We now evaluate the two integrals presented above, respectively. For convenience, we set $p_i=D-1-2 i$ and $C(D, i)=\binom{D-1}{i}(-1)^i \exp \left(\frac{\sigma^2}{2}p_i^2\right)\left\{\operatorname{erf}\left(\frac{\sigma p_i}{\sqrt{2}}\right)+1\right\}$.
\begin{equation}
\begin{aligned}
& \frac{1}{\sigma^4} \int \frac{1}{\xi(\sigma)} \exp \left(-\frac{1}{2 \sigma^2} r^2\right) r^2 \sinh ^{D-1} r d r \\
& =\sqrt{\frac{\pi}{2}} \frac{1}{2^{D-1} \xi(\sigma) \sigma}\sum_{i=0}^{D-1} C(D, i)\left(\sigma^2p_i^2+1\right) ,
\end{aligned}
\end{equation}
\begin{equation}
\label{cos_sin}
\int \cos \theta_1^2 \sin ^{D-2} \theta_1 \cdots \sin \theta_{D-2} d \theta_1 \cdots d \theta_{D-1} =\frac{\pi^{\frac{D}{2}}}{\Gamma\left(\frac{D+2}{2}\right)}.
\end{equation}
The derivation of Equation (\ref{cos_sin}) utilizes the Wallis integral.

The derivative of the normalization term of R-GD $\xi(\sigma)$ on hyperbolic space with respect to $\sigma$ is given as follows. 

\begin{equation}
\label{dim_xi}
\xi^{\prime}(\sigma)=\frac{\xi(\sigma)}{\sigma}+\frac{2 \pi^{\frac{D}{2}}}{\Gamma\left(\frac{D}{2}\right)} \sqrt{\frac{\pi}{2}} \frac{1}{2^{D-1}} \sum_{i=0}^{D-1} C(D, i) \sigma^2p_i^2.
\end{equation}
Using Equation (\ref{dim_xi}), the constant $I_\phi\left(o_h, \sigma\right)_{i, i}$ can be expressed in a simplified form as follows.

\begin{equation}
\label{easy_I}
    I_\phi\left(o_h, \sigma\right)_{i, i}= \frac{\xi^{\prime}(\sigma)}{D \sigma \xi(\sigma)}.
\end{equation}
From the above results, it follows that $I_\phi\left(o_h, \sigma\right)$ is a diagonal matrix whose diagonal entries are given by the value in Equation (\ref{easy_I}).

We now proceed to evaluate $I_{\sigma}(\sigma)$. We begin by computing the second derivative of the log-likelihood function \( L \) with respect to \( \sigma \).
\begin{equation}
\frac{\partial^2 L}{\partial \sigma^2}=-\frac{\xi(\sigma) \xi^{\prime \prime}(\sigma)-\xi^{\prime}(\sigma)^2}{\xi^2(\sigma)}-\frac{3}{\sigma^4} d^2(x, \mu) .
\end{equation}
Based on the above computation, the Fisher information $I_{\sigma}(\sigma)$ is calculated as follows.
\begin{equation}
\begin{aligned}
& I_\sigma(\sigma) \\
& =\int \frac{1}{\xi(\sigma)} \exp \left(-\frac{1}{2 \sigma^2} d^2(x, \mu)\right)\frac{\partial^2 L}{\partial \sigma^2} d \operatorname{vol}(x) \\
& =\frac{\xi(\sigma) \xi^{\prime \prime}(\sigma)-\xi^{\prime}(\sigma)^2}{\xi^2(\sigma)} \\
&+\frac{3}{\sigma\xi(\sigma)}\frac{2 \pi^{\frac{D}{2}}}{\Gamma\left(\frac{D}{2}\right)}\sqrt{\frac{\pi}{2}} \frac{1}{2^{D-1}}\sum_{i=0}^{D-1} C(D, i)\left(\sigma^2p_i^2+1\right) \\
& =\frac{\xi^{\prime \prime}(\sigma)}{\xi(\sigma)}-\left(\frac{\xi^{\prime}(\sigma)}{\xi(\sigma)}\right)^2+\frac{3}{\sigma} \frac{\xi^{\prime}(\sigma)}{\xi(\sigma)} .
\end{aligned}
\end{equation}
\bibliographystyle{IEEEtran} %文献リストのスタイルを召喚
\bibliography{biblio}
\begin{IEEEbiographynophoto}{Kota Fukuzawa}
was born in Ibaraki, Japan, in 1999. He received the M.S. degree in mathematical informatics from The University of Tokyo, Tokyo, Japan, in 2025.
Since 2025, he has been with the IOWN Integrated Innovation Center, NTT Corporation, Tokyo, Japan, where he is engaged in research on retraining of machine learning models.
\end{IEEEbiographynophoto}
\begin{IEEEbiographynophoto}{Atsushi Suzuki}
received the Ph.D. degree from the Graduate School of Information Science and Technology, The University of Tokyo, Japan, in 2020. From 2020 to 2022, Atsushi was a Lecturer with the University of Greenwich, U.K. From 2022 to 2025, Atsushi was a Lecturer with King’s College London, U.K. Since 2025, Atsushi has been an Assistant Professor with The University of Hong Kong. Atsushi’s research interests include machine learning theory using information theory, statistics, geometry, and related mathematical tools.
\end{IEEEbiographynophoto}
\begin{IEEEbiographynophoto}{Kenji Yamanishi}
(Senior Member, IEEE) received the master’s and Dr.Eng. degrees from The University of Tokyo, Tokyo, Japan, in 1987 and 1992, respectively. He joined NEC Corporation in 1987, and his final position, in 2008, as a fellow of the Internet Systems Research Laboratories. He was also the Department Head of the Data Mining Technology Center, NEC Corporation. He worked for the NEC Research Institute, USA, from 1992 to 1995, as a Visiting Scientist. In 2009, he joined The University of Tokyo, where he has been working as a Professor with the Graduate School of Information Science and Technology. His research interests include information-theoretic machine 
learning, data science, and big data analysis. He is a fellow of  the Institute of Electronics, Information and Communication Engineers (IEICE). He is the author of the book ”Learning with the Minimum Description Length Principle” (Springer).
\end{IEEEbiographynophoto}
\vspace*{219pt}
\end{document}